\documentclass{article}
\usepackage{iclr2023_conference,times}

\usepackage{microtype}

\usepackage{amsmath}
\usepackage{amssymb}
\usepackage{mathtools}
\usepackage{amsthm}
\usepackage{graphicx}
\usepackage{booktabs}
\usepackage{amsmath}
\usepackage{amssymb}
\usepackage{amsfonts}
\usepackage{multirow}
\usepackage{verbatim}
\usepackage{caption}
\usepackage{supertabular}
\usepackage{float}
\usepackage{tabularx} 
\usepackage{listings, listings-rust} 
\lstdefinelanguage{JavaScript}{
  morekeywords=[1]{break, continue, delete, else, for, function, if, in,
    new, return, this, typeof, var, void, while, with},
  morekeywords=[2]{false, null, true, boolean, number, undefined,
    Array, Boolean, Date, Math, Number, String, Object},
  morekeywords=[3]{eval, parseInt, parseFloat, escape, unescape},
  sensitive,
  morecomment=[s]{/*}{*/},
  morecomment=[l]//,
  morecomment=[s]{/**}{*/}, 
  morestring=[b]',
  morestring=[b]"
}[keywords, comments, strings]

\usepackage{enumitem}
\usepackage{xcolor}
\usepackage{xspace}
\usepackage{textcomp}
\usepackage{makecell}
\usepackage{multicol}
\usepackage{lscape} 
\usepackage{siunitx}
\usepackage{algorithm}
\usepackage{algpseudocode}

\usepackage{amssymb}
\usepackage{pifont}
\usepackage{scrextend}

\usepackage{array}
\usepackage{latexsym}
\usepackage[T1]{fontenc}
\usepackage[utf8]{inputenc}
\usepackage{microtype}
\usepackage{xurl} 
\usepackage{nicefrac}       
\usepackage{changepage}
\usepackage{xargs}
\usepackage{wrapfig,lipsum,booktabs}
\usepackage{longtable}
\usepackage{subcaption}
\usepackage{pgfplots}
\usetikzlibrary{pgfplots.groupplots}
\pgfplotsset{compat=1.3}
\usepackage{tikz}
\usetikzlibrary{patterns}
\usepackage{placeins} 

\usepackage[most]{tcolorbox}

\usepackage{hyperref}
\usepackage[capitalize,noabbrev]{cleveref}
\crefname{section}{Section}{\S\S}
\crefname{section}{Section}{\S\S}
\crefname{table}{Table}{Tables}
\crefname{figure}{Figure}{Figures}
\crefname{algorithm}{Algorithm}{}
\crefname{equation}{eq.}{}
\crefname{appendix}{Appendix}{}
\crefformat{section}{Section #2#1#3}

\definecolor{mydarkblue}{rgb}{0,0.08,0.45}
\hypersetup{
    colorlinks=true,
    linkcolor=mydarkblue,
    filecolor=blue,      
    urlcolor=mydarkblue,
    citecolor=violet
}

\newcommand{\pack}[0]{\textsc{OctoPack}}
\newcommand{\badpack}[0]{\textsc{OctoBadPack}}
\newcommand{\model}[0]{\textsc{OctoCoder}}
\newcommand{\modelx}[0]{\textsc{OctoGeeX}}
\newcommand{\modelsc}[0]{\textsc{SantaCoderPack}}
\newcommand{\data}[0]{\textsc{CommitPack}}
\newcommand{\dataft}[0]{\textsc{CommitPackFT}}
\newcommand{\eval}[0]{\textsc{HumanEvalPack}}
\newcommand{\evalf}[0]{\textsc{HumanEvalFix}}
\newcommand{\evalfd}[0]{\textsc{HumanEvalFixDocs}}

\newcommand{\evale}[0]{\textsc{HumanEvalExplain}}
\newcommand{\evals}[0]{\textsc{HumanEvalSynthesize}}

\newcommand{\papertitle}[0]{\vspace{-2mm}OctoPack: Instruction Tuning Code Large Language Models}

\DeclareSymbolFont{extraup}{U}{zavm}{m}{n}
\DeclareMathSymbol{\varheart}{\mathalpha}{extraup}{86}
\DeclareMathSymbol{\vardiamond}{\mathalpha}{extraup}{87}

\iclrfinalcopy

\usepackage[capitalize,noabbrev]{cleveref}

\theoremstyle{plain}

\theoremstyle{definition}

\theoremstyle{remark}

\usepackage[loose]{minitoc} 

\noptcrule

\makeatletter
\AtBeginDocument{%
  \renewcommand{\sectionautorefname}{\S\@gobble}

  \renewcommand{\subsectionautorefname}{\S\@gobble}  
}
\makeatother

\title{\raggedright \papertitle}

\author{
\vspace{2.5mm} 
\hspace{-6.5mm}
        Niklas Muennighoff\hspace{6.5mm}
        Qian Liu\hspace{5.5mm}
        Armel Zebaze\hspace{5.5mm}
        Qinkai Zheng\hspace{5.5mm}
        Binyuan Hui
    \\
    \vspace{2.5mm}
    \textbf{\hspace{-5.5mm}
        Terry Yue Zhuo\hspace{4mm}
        Swayam Singh\hspace{4mm}
        Xiangru Tang\hspace{4mm}
        Leandro von Werra\hspace{4mm}
        Shayne Longpre
    }
    \\
    \hspace{-5.5mm}
    {\tt \href{mailto:n.muennighoff@gmail.com}{n.muennighoff@gmail.com}}
}

\fancypagestyle{firstpage}{
  \lhead{\pack{}: Instruction Tuning Code Large Language Models}
  \rhead{\normalsize{\today}}
}

\begin{document}

\doparttoc 
\faketableofcontents 

\maketitle
\thispagestyle{empty}

\begin{tikzpicture}[remember picture,overlay,shift={(current page.north west)}]
\node[anchor=north west,xshift=17cm,yshift=-1.5cm]{\scalebox{0.7}[0.7]{\includegraphics[width=5.5cm]{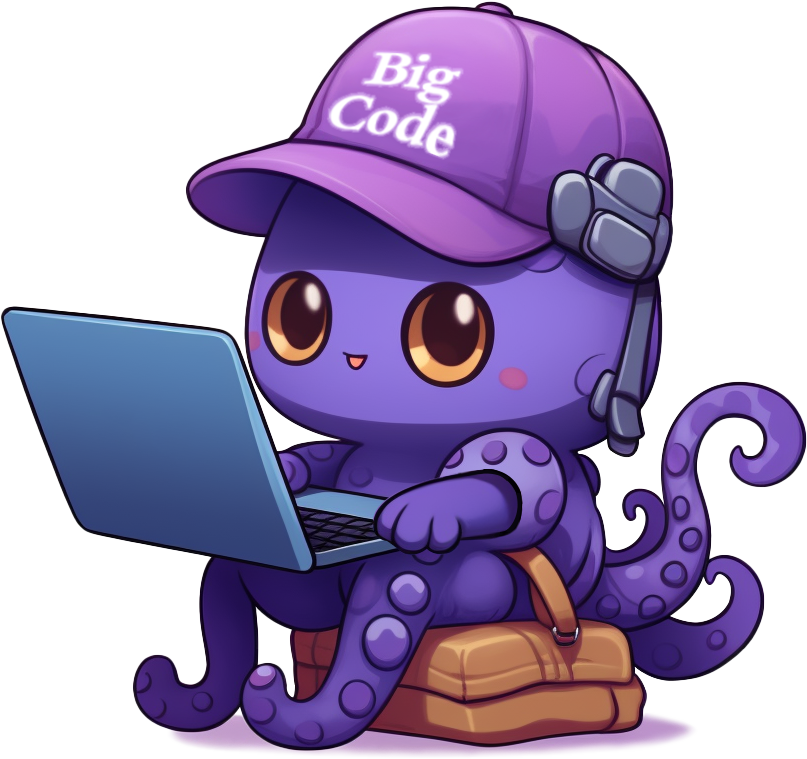}}};
\end{tikzpicture}

\vspace{-14mm}
\begin{abstract}

Finetuning large language models (LLMs) on instructions leads to vast performance improvements on natural language tasks. We apply instruction tuning using code, leveraging the natural structure of Git commits, which pair code changes with human instructions. We compile \data{}: 4 terabytes of Git commits across 350 programming languages. We benchmark \data{} against other natural and synthetic code instructions (xP3x, Self-Instruct, OASST) on the 16B parameter StarCoder model, and achieve state-of-the-art performance among models not trained on OpenAI outputs, on the HumanEval Python benchmark (46.2\% pass@$1$). We further introduce \eval{}, expanding the HumanEval benchmark to a total of 3 coding tasks (Code Repair, Code Explanation, Code Synthesis) across 6 languages (Python, JavaScript, Java, Go, C++, Rust). Our models, \model{} and \modelx{}, achieve the best performance across \eval{} among all permissive models, demonstrating \data{}'s benefits in generalizing to a wider set of languages and natural coding tasks. Code, models and data are freely available at \url{https://github.com/bigcode-project/octopack}.



\end{abstract}

\vspace{-2mm}
\begin{figure}[htbp]
    \centering
    \includegraphics[width=\linewidth]{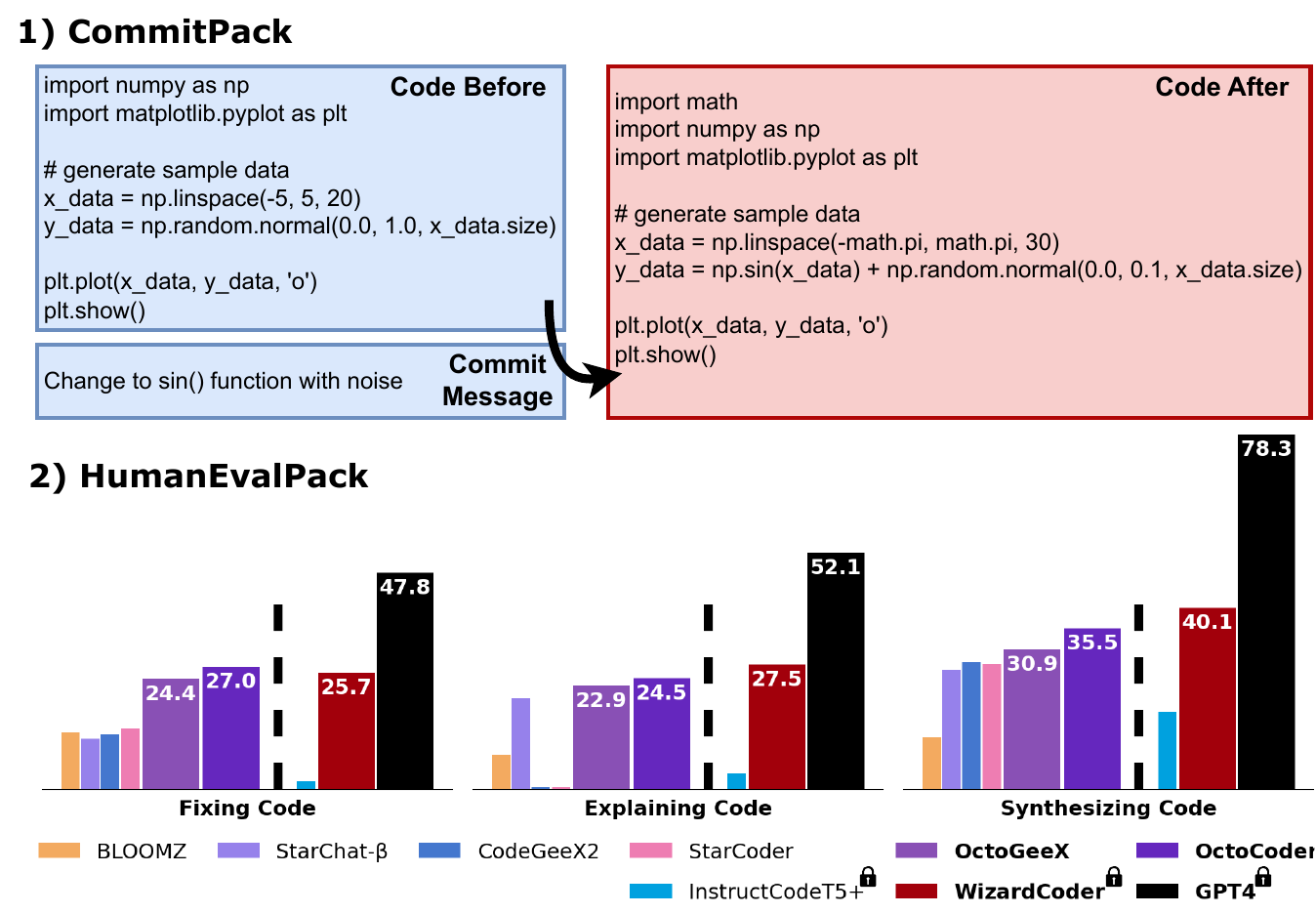}
    \caption{\textbf{\pack{} Overview.} \textit{1)} Sample from our 4TB dataset, \data{}. \textit{2)}~Performance of \model{}, \modelx{} and other code models including non-permissive ones (WizardCoder, GPT-4) on \eval{} spanning 3 coding tasks and 6 programming languages.}
    \label{fig:main}
\end{figure}

\newpage

\section{Introduction}
\label{sec:intro}

Finetuning large language models (LLMs) on a variety of language tasks explained via instructions (instruction tuning) has been shown to improve model usability and general performance~\citep{wei2021finetuned,sanh2022multitask,min2021metaicl,ouyang2022training}. The instruction tuning paradigm has also proven successful for models trained on visual~\citep{liu2023visual,li2023otter}, audio~\citep{zhang2023video} and multilingual~\citep{muennighoff2022crosslingual,wang2022super} data.

In this work, we instruction tune LLMs on the coding modality. While Code LLMs can already be indirectly instructed to generate desired code using code comments, this procedure is brittle and does not work when the desired output is natural language, such as explaining code. Explicit instructing tuning of Code LLMs may improve their steerability and enable their application to more tasks. Concurrently to our work, three instruction tuned Code LLMs have been proposed: PanGu-Coder2~\citep{shen2023pangucoder2}, WizardCoder~\citep{luo2023wizardcoder} and InstructCodeT5+~\citep{wang2023codet5+}. These models rely on more capable and closed models from the OpenAI API\footnote{\url{https://openai.com/blog/openai-api}} to create their instruction training data. This approach is problematic as \textbf{(1)} closed-source APIs keep changing and have unpredictable availability~\citep{pozzobon2023challenges,chen2023chatgpts}, \textbf{(2)} it relies on the assumption that a more capable model exists \textbf{(3)} it can reinforce model hallucination~\citep{gudibande2023false} and \textbf{(4)}, depending on legal interpretation, OpenAI's terms of use\footnote{\url{https://openai.com/policies/terms-of-use}} forbid such models: ``...You may not...use output from the Services to develop models that compete with OpenAI...''. Thus, we consider models trained on OpenAI outputs not usable for commercial purposes in practice and classify them as non-permissive in this work.

We focus on more permissively licensed data and avoid using a closed-source model to generate synthetic data. We benchmark four popular sources of code instruction data: \textbf{(1)} xP3x~\citep{muennighoff2022crosslingual}, which contains data from common code benchmarks, \textbf{(2)} Self-Instruct~\citep{wang2022self} data we create using a permissive Code LLM, \textbf{(3)} OASST~\citep{kopf2023openassistant}, which contains mostly natural language data and few code examples and \textbf{(4)} \data{}, our new 4TB dataset of Git commits. Instruction tuning's primary purpose is to expand models' generalization abilities to a wide variety of tasks and settings. Thus, we extend the code synthesis benchmark, HumanEval~\citep{chen2021evaluating,zheng2023codegeex}, to create \eval{}: A code benchmark covering code synthesis, code repair, and code explanation across six programming languages.

Instruction tuning StarCoder~\citep{li2023starcoder} on a filtered variant of \data{} and OASST leads to our best model, \model{}, which surpasses all other openly licensed models (\autoref{fig:main}), but falls short of the much larger GPT-4~\citep{openai2023gpt4}. GPT-4 is close to maximum performance on the code synthesis variant, notably with a pass@$1$ score of 86.6\% on Python HumanEval. However, it performs significantly worse on the code fixing and explanation variants of \eval{}, which we introduce. This suggests that the original HumanEval benchmark may soon cease to be useful due to models reaching close to the maximum performance. Our more challenging evaluation variants provide room for future LLMs to improve on the performance of the current state-of-the-art.

In summary, we contribute:
\begin{itemize}
\item \data{} and \dataft{}: 4TB of permissively licensed code commits across 350 programming languages for pretraining and a filtered 2GB variant containing high-quality code instructions used for finetuning
\item \eval{}: A benchmark for Code LLM generalization, spanning three scenarios (Code Repair, Code Explanation, Code Synthesis) and 6 programming languages (Python, JavaScript, Java, Go, C++, Rust)
\item \model{} and \modelx{}: The best permissive Code LLMs
\end{itemize}

\section{\data{}: Code Instruction Data}
\label{sec:data}

\begin{figure*}[!ht]
    \centering
    {{\includegraphics[width=\textwidth]{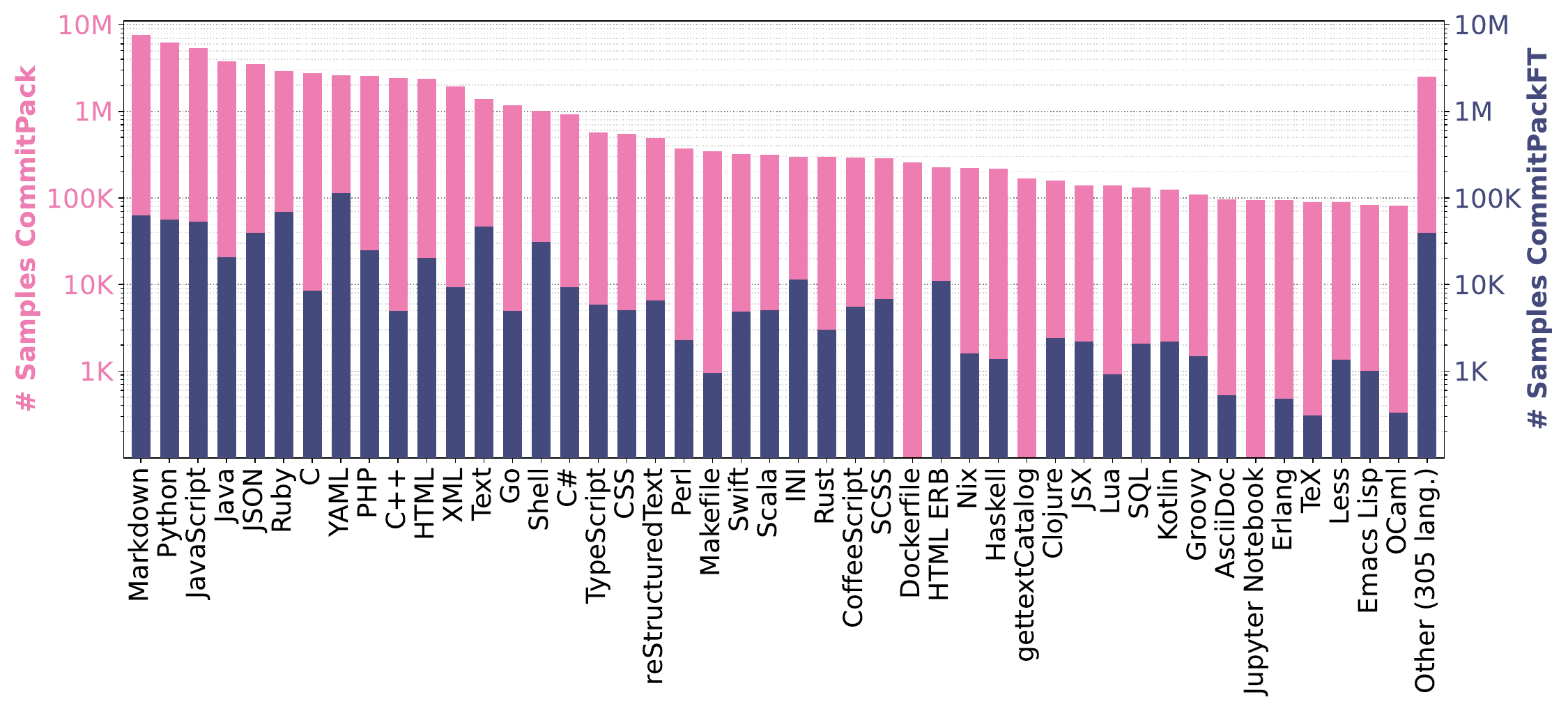}}}
    \qquad    
    {{\includegraphics[width=0.9\textwidth]{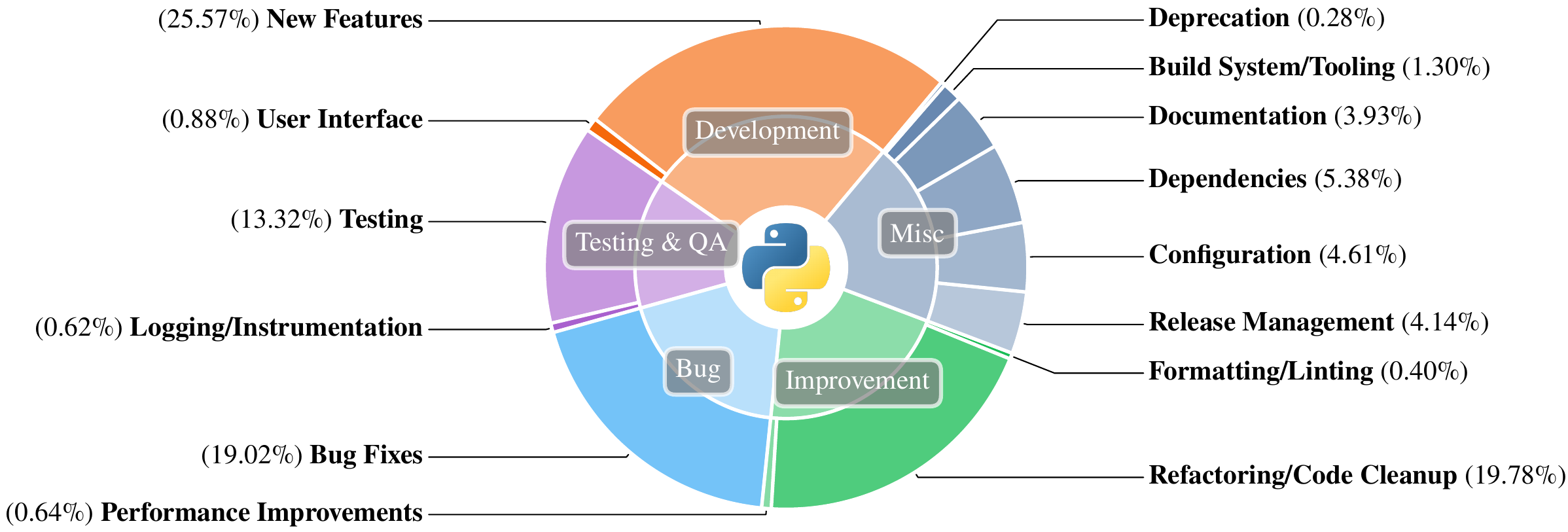}}}
    \caption{\textbf{Overview of \data{} and \dataft{}.} \textit{Top:} Language distribution of the full commit data (\data{}) and the variant filtered for high-quality instructions (\dataft{}). See \autoref{sec:dataapp} for the full distribution. \textit{Bottom:} Task distribution of commits on the Python subset of \dataft{} (59K samples) according to GPT-4.}
    \label{fig:data}
\end{figure*}

\begin{table}[htbp]
    \centering
    \resizebox{1\textwidth}{!}{
    \begin{tabular}{l|ccc|ccc}
    \toprule
    & \multicolumn{3}{c|}{Base dataset} & \multicolumn{3}{c}{Subset} \\
    \midrule
    Dataset ($\downarrow$) & Lang. & Samples & Code fraction & Lang. & Samples & Code fraction \\
    \midrule
    xP3x & 8 & 532,107,156 & 0.67\% & 8 & 5,000 & 100\% \\    
    StarCoder Self-Instruct & 12 & 5,003 & 100\% & 12 & 5,003 & 100\% \\
    OASST & 49 & 161,443 & 0.9\% & 28 & 8,587 & 2.5\% \\
    \dataft{} & 277 & 742,273 & 100\% & 6 & 5,000 & 100\% \\    
    \bottomrule
    \end{tabular}
    }
    \caption{\textbf{Statistics of code instruction data we consider.} We display the number of programming languages, total samples, and fraction of samples that contain code for permissive instruction datasets. For finetuning on these datasets, we use small subsets with around 5,000 samples each.}
    \label{tab:ablationsdata}
\end{table}

Prior work has shown that models can generalize to languages included in pretraining, but absent during instruction tuning ~\citep{muennighoff2022crosslingual}. However, they also show that including such languages during instruction tuning boosts their performance further. We hypothesize that code data exhibits the same behavior. To improve performance on code-related tasks, we thus construct a code instruction dataset leveraging the natural structure of Git commits.

\paragraph{\data{}} To create the dataset, we use commit metadata from the GitHub action dump on Google BigQuery.\footnote{\url{https://www.gharchive.org/}} We apply quality filters, filter for commercially friendly licenses, and discard commits that affect more than a single file to ensure commit messages are very specific and to avoid additional complexity from dealing with multiple files. We use the filtered metadata to scrape the affected code files prior to and after the commit from GitHub. This leads to almost 4 terabytes of data covering 350 programming languages (\data{}). As instruction tuning does not require so much data~\citep{zhou2023lima,touvron2023llama2}, we apply several strict filters to reduce the dataset to 2 gigabytes and 277 languages (\dataft{}). These include filtering for samples where the commit message has specific words in uppercase imperative form at the start (e.g. "Verify ..."), consists of multiple words, and does not contain external references. All filters are detailed in~\autoref{sec:datacreation}. \autoref{fig:data} depicts the distribution of both datasets and the tasks contained in \dataft{}. For instruction tuning our models, we select 5,000 random samples from \dataft{} across the 6 programming languages that we evaluate on. In~\autoref{sec:pretraining}, we also experiment with pretraining on the entirety of \data{}.

\paragraph{Alternatives} We consider three additional datasets for instruction tuning presented in~\autoref{tab:ablationsdata}. 
\textbf{xP3x}: xP3x is a large-scale collection of multilingual instruction data with around 532 million samples~\citep{muennighoff2022crosslingual}. We focus only on the code subset of xP3x, excluding NeuralCodeSearch~\citep{li2019neural} which is not licensed permissively, and select 5,000 samples. \\
\textbf{Self-Instruct}: Using the Self-Instruct method~\citep{wang2022selfinstruct} and the StarCoder model~\citep{li2023starcoder}, we create 5,003 synthetic instructions and corresponding answers.\\ 
\textbf{OASST}: OASST is a diverse dataset of multi-turn chat dialogues~\citep{kopf2023openassistant}. Only a few of the dialogues contain code. We reuse a filtered variant from prior work~\citep{dettmers2023qlora} and additionally filter out moralizing assistant answers (\autoref{sec:datacreation}) leading to 8,587 samples.

\section{\eval{}: Evaluating Instruction Tuned Code Models}
\label{sec:eval}

\begin{figure}[htbp]
    \centering
    \includegraphics[width=\linewidth]{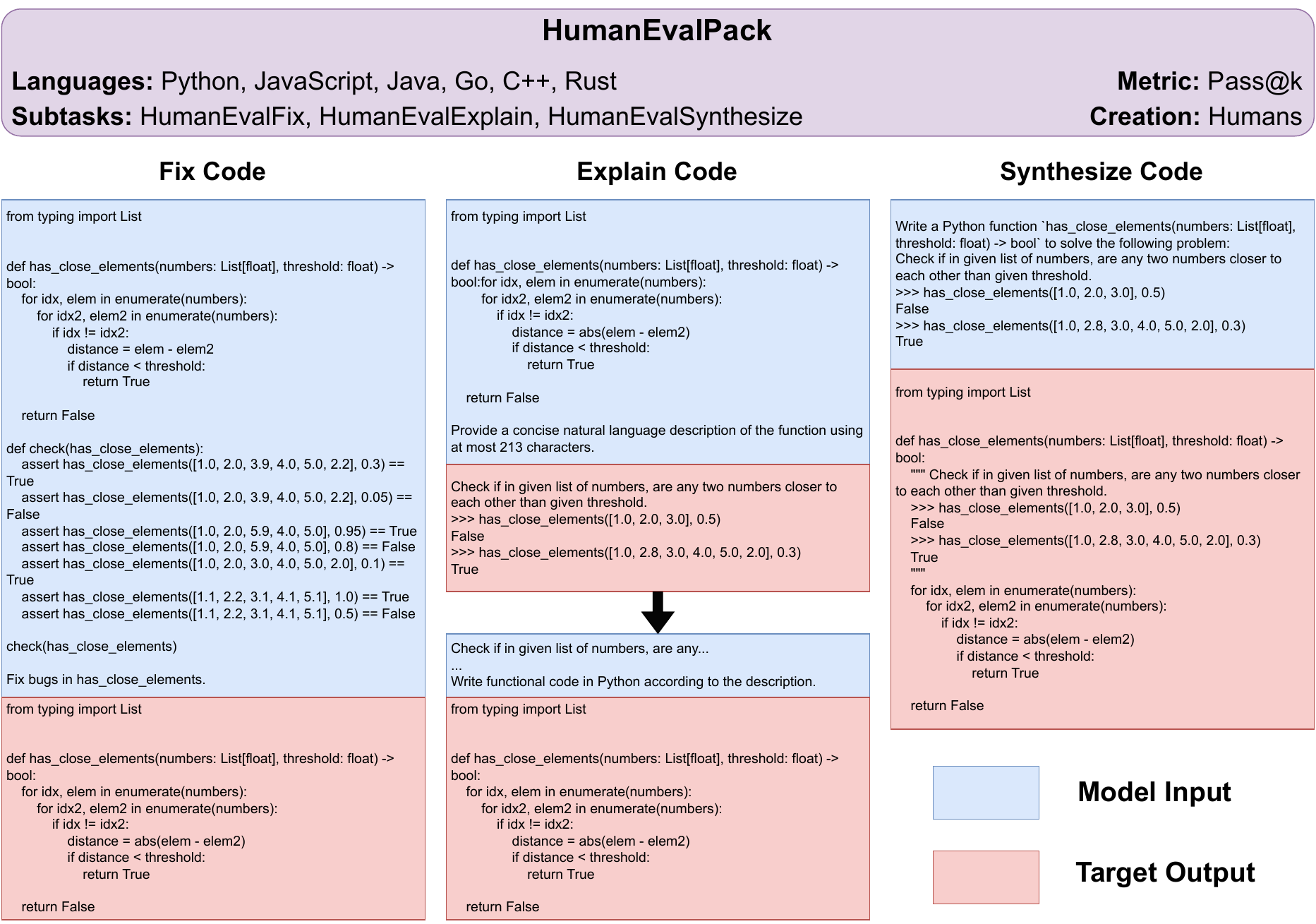}
    \caption{\textbf{\eval{} overview.} The first HumanEval problem is depicted across the three scenarios for Python. The bug for \evalf{} consists of a missing "abs" statement.}
    \label{fig:eval}
\end{figure}

When instruction tuning LLMs using natural language (NL) data, the input is an NL instruction with optional NL context and the target output is the NL answer to the task~\citep{wei2021finetuned}. When instruction tuning with code (C) data, code may either appear \textit{only in the input} alongside the NL instruction (NL+C$\rightarrow$NL, e.g. code explanation), \textit{only in the output} (NL$\rightarrow$C, e.g. code synthesis), or in \textit{both input and output} (NL+C$\rightarrow$C, e.g. code modifications like bug fixing). While prior benchmarks commonly only cover variants of code synthesis, users may want to use models in all three scenarios. Thus, we expand the code synthesis benchmark HumanEval~\citep{chen2021evaluating,zheng2023codegeex} to cover all three input-output combinations for six languages (\autoref{fig:eval}).

\paragraph{\evalf{} (NL+C$\rightarrow$C)} Given an incorrect code function with a subtle bug and accompanying unit tests, the model is tasked to fix the function. We manually add a bug to each of the 164 HumanEval solutions across all 6 languages (984 total bugs). For a given sample, the bugs are as similar as possible across the 6 languages enabling meaningful comparison of scores across languages. Bugs are written such that the code still runs but produces an incorrect result leading to at least one unit test failing. Bug statistics and examples are in \autoref{sec:bugs}. We also evaluate an easier variant of this task where instead of unit tests, models are provided with the correct function docstring as the source of truth to fix bugs, see \autoref{sec:docstrings}.

\paragraph{\evale{} (NL+C$\rightarrow$NL)} Given a correct code function, the model is tasked to generate an explanation of the code. Subsequently, the same model is tasked to regenerate the code given only its own explanation. The second step allows us to score this task via code execution and measure pass@$k$~\citep{chen2021evaluating} instead of evaluating the explanation itself using heuristic-based metrics like BLEU~\citep{papineni2002bleu} or ROUGE~\citep{lin2004rouge} which have major limitations~\citep{reiter2018structured,schluter2017limits,eghbali2022crystalbleu,zhou2023codebertscore}. To prevent models from copying the solution into the description, we remove any solution overlap of at least 20 characters from the description. We further enforce a character length limit on the model-generated explanation equivalent to the length of the docstring describing the function. This limit is specified in the prompt for the model. Note that the function docstring itself is never provided to the model for this task.

\paragraph{\evals{} (NL$\rightarrow$C)} Given a natural language docstring or comment describing the desired code, the model is tasked to synthesize the correct code. This task corresponds to the original HumanEval benchmark~\citep{chen2021evaluating}. For instruction tuned models, we add an explicit instruction to the input explaining what the model should do. For models that have only gone through language model pretraining, we follow \cite{chen2021evaluating} and provide the model with the function header and docstring to evaluate its completion of the function.

For all tasks we execute the code generations to compute performance using the pass@$k$ metric~\citep{chen2021evaluating}: a problem is considered solved if any of $k$ code generations passes every test case. We focus on the simplest version of pass@$k$, which is pass@$1$: the likelihood that the model solves a problem in a single attempt. Like \citet{chen2021evaluating}, we use a sampling temperature of $0.2$ and $top_p=0.95$ to estimate pass@$1$. We generate $n=20$~samples, which is enough to get reliable pass@$1$ estimates~\citep{li2023starcoder}. For GPT-4, we generate $n=1$~samples. Using $n=1$ instead of $n=20$ for GPT-4 only changed scores from 75.0\% to 75.2\% pass@$1$ on \evals{} Python while providing 20x cost savings.

Python HumanEval is the most widely used code benchmark and many training datasets have already been decontaminated for it~\citep {kocetkov2022stack}. By manually extending HumanEval, we ensure existing decontamination remains valid to enable fair evaluation. However, this may not hold for all models (e.g. GPT-4), thus results should be interpreted carefully.

\section{\model{}: Best Commercially Licensed Code LLM}
\label{sec:model}

\subsection{Ablating instruction data choices}
\label{sec:ablate}

\begin{figure}[htbp]
    \centering
    \includegraphics[width=\linewidth]{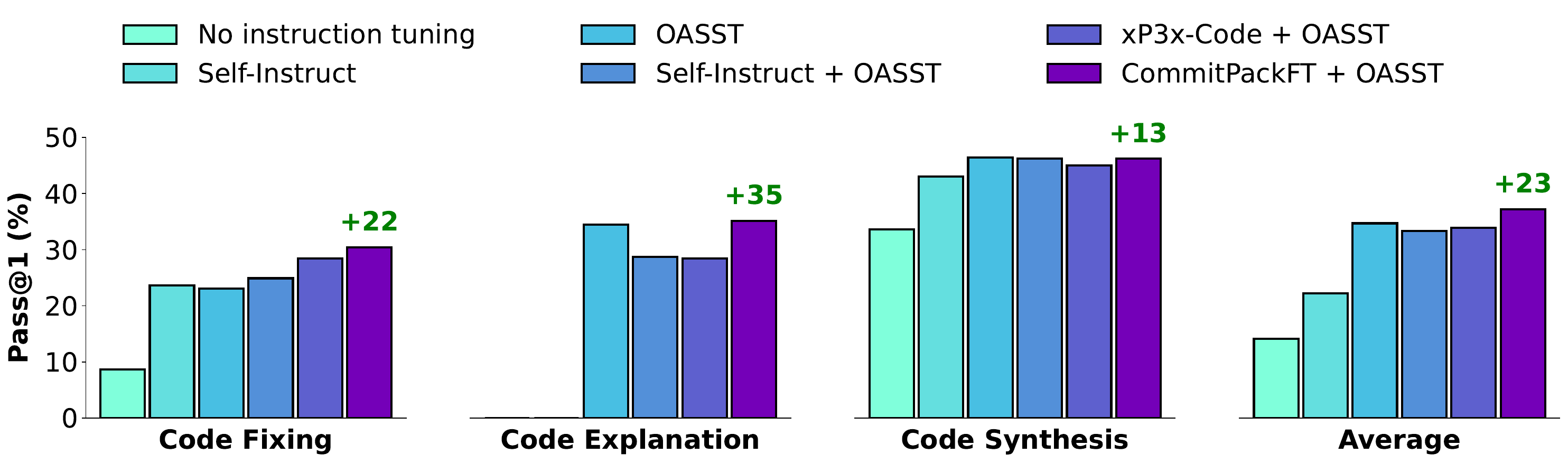}
    \caption{\textbf{Comparing permissively licensed instruction datasets by instruction tuning StarCoder.} Models are evaluated on the Python subset of \eval{}.}
    \label{fig:ablations}
\end{figure}

We instruction tune the pretrained StarCoder model~\citep{li2023starcoder} on different combinations of our instruction datasets (\autoref{sec:data}). We evaluate all models on the Python subset of \eval{} as depicted in~\autoref{fig:ablations}. Similar to prior work~\citep{alpaca}, we format all instructions into a consistent schema to distinguish question and answer (see \autoref{fig:octocoder}).

\paragraph{\dataft{} enables CodeLLMs to fix bugs} \dataft{} is critical for the performance boost on code repair (\evalf{}), where instruction tuning on only OASST or other variants results in a significantly lower score. This is likely due to \dataft{} including around 20\% of bug fixes among other code-related tasks (\autoref{fig:data}).

\paragraph{Importance of samples with natural language targets} The pretrained StarCoder model, as well as the Self-Instruct variant, perform poorly on code explanation (\evale{}). This is because both models are only conditioned to write code instead of natural language. We find that to perform well at explaining code, it is necessary to include samples with natural language as the target output during instruction tuning. Only relying on data with code as the target, such as the Self-Instruct data, will lead to models always outputting code even if the question requires a natural language output. Thus, we mix all other ablations with OASST, which contains many natural language targets. While the xP3x subset also contains samples with natural language output, many of its target outputs are short, which leads to models with a bias for short answers. This is impractical for the explanation task leading to the comparatively low score of mixing xP3x with OASST.

\paragraph{\dataft{}+OASST yields best performance} All instruction datasets provide similar boosts for code synthesis (\evals{}), which has been the focus of all prior work on code instruction models~\citep{wang2023codet5+,luo2023wizardcoder,muennighoff2022crosslingual}. We achieve the best average score by instruction tuning on \dataft{} mixed with our filtered OASST data yielding an absolute 23\% improvement over StarCoder. Thus, we select \dataft{}+OASST for our final model dubbed \model{}. Using the same data, we also instruction tune the 6 billion parameter CodeGeeX2~\citep{zheng2023codegeex} to create \modelx{}. Training hyperparameters for both models are in~\autoref{sec:hyperparameters}.

\subsection{Comparing with other Models}
\label{sec:comp}

\begin{table*}[htbp]
    \centering
    \begin{tabular}{l|cccccc|c}
    \toprule
    \textbf{Model ($\downarrow$)} & \textbf{Python} & \textbf{JavaScript} & \textbf{Java} & \textbf{Go} & \textbf{C++} & \textbf{Rust} & \textbf{Avg.} \\
    \midrule
    \midrule     
    \multicolumn{8}{c}{} \\
    \multicolumn{8}{c}{\textbf{\evalf{}}} \\
    \multicolumn{8}{c}{} \\
    \midrule
    \multicolumn{8}{c}{Non-permissive models} \\
    \midrule
    InstructCodeT5+$^\dagger$ & 2.7 & 1.2 & 4.3 & 2.1 & 0.2 & 0.5 & 1.8\\    
    WizardCoder$^\dagger$ & 31.8 & 29.5 & 30.7 & 30.4 & 18.7 & 13.0 & 25.7 \\
    GPT-4 & 47.0 & 48.2 & 50.0 & 50.6 & 47.6 & 43.3 & \underline{47.8} \\
    \midrule
    \multicolumn{8}{c}{Permissive models} \\
    \midrule
    BLOOMZ & 16.6 & 15.5 & 15.2 & 16.4 & 6.7 & 5.7 & 12.5\\
    StarChat-$\beta$ & 18.1 & 18.1 & 24.1 & 18.1 & 8.2 & 3.6 & 11.2 \\
    CodeGeeX2$^{*}$ & 15.9 & 14.7 & 18.0 & 13.6 & 4.3 & 6.1 & 12.1 \\
    StarCoder & 8.7 & 15.7 & 13.3 & 20.1 & 15.6 & 6.7 & 13.4 \\
    \modelx{}$^{*}$ & 28.1 & 27.7 & 30.4 & 27.6 & 22.9 & 9.6 & 24.4 \\
    \model{} & \textbf{30.4} & \textbf{28.4} & \textbf{30.6} & \textbf{30.2} & \textbf{26.1} & \textbf{16.5} & \textbf{27.0} \\
    \midrule
    \midrule    
    \multicolumn{8}{c}{} \\    
    \multicolumn{8}{c}{\textbf{\evale{}}} \\
    \multicolumn{8}{c}{} \\
    \midrule    
    \multicolumn{8}{c}{Non-permissive models} \\
    \midrule
    InstructCodeT5+$^\dagger$ & 20.8 & 0.0 & 0.0 & 0.0 & 0.1 & 0.0 & 3.5 \\    
    WizardCoder$^\dagger$ & 32.5 & 33.0 & 27.4 & 26.7 & 28.2 & 16.9 & 27.5 \\
    GPT-4 & 64.6 & 57.3 & 51.2 & 58.5 & 38.4 & 42.7 & \underline{52.1} \\
    \midrule
    \multicolumn{8}{c}{Permissive models} \\
    \midrule
    BLOOMZ & 14.7 & 8.8 & 12.1 & 8.5 & 0.6 & 0.0 & 7.5 \\
    StarChat-$\beta$ & 25.4 & 21.5 & 24.5 & 18.4 & 17.6 & 13.2 & 20.1 \\ 
    CodeGeeX2$^{*}$ & 0.0 & 0.0 & 0.0 & 0.0 & 0.0 & 0.0 & 0.0 \\
    StarCoder & 0.0 & 0.0 & 0.0 & 0.0 & 0.0 & 0.0 & 0.0 \\
    \modelx{}$^{*}$ & 30.4 & 24.0 & 24.7 & \textbf{21.7} & 21.0 & \textbf{15.9} & 22.9 \\
    \model{} & \textbf{35.1} & \textbf{24.5} & \textbf{27.3} & 21.1 & \textbf{24.1} & 14.8 & \textbf{24.5} \\
    \midrule
    \midrule    
    \multicolumn{8}{c}{} \\
    \multicolumn{8}{c}{\textbf{\evals{}}} \\
    \multicolumn{8}{c}{} \\
    \midrule
    \multicolumn{8}{c}{Non-permissive models} \\
    \midrule
    InstructCodeT5+$^\dagger$ & 37.0 & 18.9 & 17.4 & 9.5 & 19.8 & 0.3 & 17.1 \\    
    WizardCoder$^\dagger$ & 57.3 & 49.5 & 36.1 & 36.4 & 40.9 & 20.2 & 40.1 \\
    GPT-4 & 86.6 & 82.9 & 81.7 & 72.6 & 78.7 & 67.1 & \underline{78.3} \\
    \midrule
    \multicolumn{8}{c}{Permissive models} \\
    \midrule
    BLOOMZ & 15.6 & 14.8 & 18.4 & 8.4 & 6.5 & 5.5 & 11.5 \\
    StarChat-$\beta$ & 33.5 & 31.4 & 26.7 & 25.5 & 26.6 & 14.0 & 26.3 \\
    CodeGeeX2$^{*}$ & 35.9 & 32.2 & 30.8 & 22.5 & 29.3 & 18.1 & 28.1 \\    
    StarCoder & 33.6 & 30.8 & 30.2 & 17.6 & 31.6 & 21.8 & 27.6 \\
    \modelx{}$^{*}$ & 44.7 & 33.8 & 36.9 & 21.9 & 32.3 & 15.7 & 30.9 \\
    \model{} & \textbf{46.2} & \textbf{39.2} & \textbf{38.2} & \textbf{30.4} & \textbf{35.6} & \textbf{23.4} & \textbf{35.5} \\
    \bottomrule
    \end{tabular}
    \caption{
        \textbf{Zero-shot pass@$1$ (\%) performance across \eval{}.} InstructCodeT5+, WizardCoder, StarChat-$\beta$, StarCoder and \model{} have 16B parameters. CodeGeeX2 and \modelx{} have 6B parameters. BLOOMZ has 176B parameters. In this work, we call models "permissive" if weights are freely accessible and usable for commercial purposes.  $^{*}$: Commercial license available after submitting a form. $\dagger$: Trained on data that may not be used ``to develop models that compete with OpenAI'' thus we classify them as non-permissive in this work (see \autoref{sec:intro}).
    }
    \label{tab:humanevalpack}
\end{table*}

We benchmark \model{} and \modelx{} with state-of-the-art Code LLMs on \eval{} in \autoref{tab:humanevalpack}. For all models, we use the prompt put forward by the model creators if applicable or else a simple intuitive prompt, see \autoref{sec:prompts}. 

\vspace{-1mm}
\paragraph{\model{} performs best among permissive models} \model{} has the highest average score across all three evaluation scenarios among all permissive models. With just 6 billion parameters, \modelx{} is the smallest model benchmarked, but still outperforms \emph{all} prior permissive Code LLMs. GPT-4~\citep{openai2023gpt4} performs best among all models benchmarked with a significant margin. However, GPT-4 is closed-source and likely much larger than all other models evaluated.

\vspace{-1mm}
\paragraph{Instruction tuning generalizes to unseen programming languages} Trained primarily on natural language, not code, BLOOMZ~\citep{muennighoff2022crosslingual} performs worse than other models despite having 176 billion parameters. Go and Rust are not contained in BLOOMZ's instruction data, yet it performs much better than the random baseline of 0.0 for these two languages across most tasks. 
This confirms our hypothesis that models are capable of generalizing instructions to programming languages only seen at pretraining, similar to crosslingual generalization for natural languages~\citep{muennighoff2022crosslingual}. 
To improve programming language generalization further, we tune \model{} and \modelx{} on many languages from \dataft{}, and this generalization improvement is reflected in the performance on \eval{}'s new languages.

\vspace{-1mm}
\paragraph{Pretraining weight correlates with programming language performance after instruction tuning} Prior work has shown that the performance on natural languages after instruction tuning is correlated with the weight of these languages during pretraining~\citep{muennighoff2022crosslingual}. The more weight during pretraining, the better the performance after instruction tuning. We find the same to be the case for programming languages. Python, Java, and JavaScript collectively make up around 30\% of the pretraining data of StarCoder~\citep{li2023starcoder}. After instruction tuning StarCoder to produce \model{}, we see the best performance among these three languages, especially for \evals{}. \model{} performs weakest on Rust, which is the lowest resource language of StarCoder among the languages we benchmark (1.2\% of pretraining data).

\vspace{-1mm}
\paragraph{Models struggle with small targeted changes} \evalf{} is the most challenging task for most models. They commonly regenerate the buggy function without making any change (e.g. WizardCoder in \autoref{fig:wizardcoderpy}) or they introduce new bugs (e.g. GPT-4 in \autoref{fig:gpt4pyaddbugs}). We analyze model performance by bug type in \autoref{sec:modelbugs} and find bugs that require removing excess code are the most challenging. \model{} performs comparatively well across all languages. Instruction tuning on \dataft{} has likely taught \model{} to make small, targeted changes to fix bugs.

\vspace{-1mm}
\paragraph{Models struggle switching between code and text} Some models fail at \evale{}, as they do not generate natural language explanations. We manually inspect explanations for the first ten samples of the Python split and disqualify a model if none of them are explanations. This is the case for StarCoder and CodeGeeX2, which generate code instead of natural language explanations. BLOOMZ and InstructCodeT5+ also occasionally generate code. Other models exclusively generate natural language explanations, not containing any code for inspected samples.

\vspace{-1mm}
\paragraph{Models struggle adhering to a specified output length} \evale{} instructs models to fit their explanation within a given character limit (\autoref{sec:eval}). Current models appear to have no understanding of how many characters they are generating. They commonly write very short and thus underspecified explanations (e.g. BLOOMZ in \autoref{fig:bloomzjs}) or excessively long explanations that end up being cut off (e.g. StarChat-$\beta$ in \autoref{fig:starchatbetacpp}). Future work could investigate how to enable models to be aware of their generated output length to improve \evale{} performance.

\vspace{-1mm}
\paragraph{HumanEval code synthesis is close to saturation} Pure code synthesis on \evals{} is the easiest task for all models. With a pass rate of 86.6\% for a single solution, GPT-4 is close to fully saturating the Python subset. GPT-4 was originally found to score 67\% on Python HumanEval~\citep{openai2023gpt4} and 81\% in later work~\citep{bubeck2023sparks}. Our score for GPT-4 is significantly higher, possibly due to improvements made to the API by OpenAI, contamination of HumanEval in GPT-4 training, or slightly different prompting and evaluation. An example of our prompt is depicted in~\autoref{fig:eval} (right). We perform very careful evaluation to ensure every generation is correctly processed. We reproduce the HumanEval score of WizardCoder~\citep{luo2023wizardcoder,xu2023wizardlm} and find it to also perform well across other languages. For BLOOMZ and InstructCodeT5+ our evaluation leads to a higher Python score than they reported, likely because of our more careful processing of generations. \model{} has the highest performance for every language among permissively licensed models. With a pass@$1$ of 46.2\% on the original Python split, \model{} improves by a relative 38\% over its base model, StarCoder. 


\section{Related Work}
\label{sec:rel}

\subsection{Code Models}

There has been extensive work on code models tailored to a specific coding task, such as code summarization~\citep{iyer2016summarizing,ahmad2020transformer,zhang2022survey,shi2022evaluation} or code editing~\citep{drain2021deepdebug,zhang2022coditt5,he2022distribution,zhang2022repairing,wei2023coeditor,prenner2023runbugrun,fakhoury2023towards,skreta2023errors} (also see work on edit models more generally \citep{reid2022learning,schick2022peer,dwivedi2022editeval,raheja2023coedit}). These works use task-specific heuristics that limit the applicability of their methods to other tasks. In contrast, we aim to build models applicable to all kinds of tasks related to code and beyond. 

Through large-scale pretraining more generally applicable code models have been developed~\citep{nijkamp2022codegen,nijkamp2023codegen2,xu2022systematic,christopoulou2022pangu,gunasekar2023textbooks,li2023starcoder,bui2023codetf,scao2022bloom,scao2022language}. However, these models only continue code making them hard to use for tasks such as explaining code with natural language (\evale{}). Teaching them to follow human instructions is critical to make them applicable to diverse tasks.

\subsection{Instruction Models}

Training models to follow instructions has led to new capabilities in text \citep{ouyang2022training, wang2022super, chung2022scaling} and visual modalities \citep{xu2023multiinstruct, openai2023gpt4}. Prior work has shown its benefits for traditional language tasks \citep{wei2021finetuned, longpre2023flan, iyer2022optiml}, multilingual tasks \citep{muennighoff2022crosslingual,muennighoff2024generative,yong2022bloom+,ustun2024aya}, and dialog \citep{kopf2023openassistant,bai2022training,gangulired}. For coding applications, PanGu-Coder2~\citep{shen2023pangucoder2}, WizardCoder~\citep{luo2023wizardcoder} and InstructCodeT5+~\citep{wang2023codet5+} are recent models trained with coding instructions. However, they all use the CodeAlpaca dataset \citep{codealpaca}, which is synthetically generated from OpenAI models. Using data from powerful closed-source models provides a strong advantage, but limits the model use and has other limitations highlighted in~\autoref{sec:intro}. CoEditor \citep{wei2023coeditor} proposes an ``auto-editing'' task, trained on 1650 python commit history repositories. Our work expands this to more general coding tasks via instructions, more languages, and orders of magnitude more commit data.

\subsection{Code Benchmarks}

Many code synthesis benchmarks have been proposed~\citep{wang2022execution,wang2022mconala,yu2023codereval,lai2023ds,du2023classeval}. HumanEval~\citep{chen2021evaluating,liu2023your} has emerged as the standard for this task. Prior work has extended HumanEval to new programming languages via automatic translation mechanisms~\citep{athiwaratkun2022multi,cassano2023multipl,orlanski2023measuring}. These approaches are error-prone and only translate tests, not the actual solutions, which are needed for tasks like code explanation. Thus, we rely only on humans to create all parts of \eval{} including test cases, correct solutions, buggy solutions, and other metadata across 6 languages.

Code repair is commonly evaluated on Quixbugs~\citep{lin2017quixbugs,prenner2021automatic,ye2021comprehensive,xia2023conversational,jiang2023impact,sobania2023analysis} or Python bugs~\citep{he2022distribution,bradley2023diffmodels}. The latter does not support code execution, which limits its utility. While Quixbugs supports execution with unit tests, it only contains 40 samples in Python and Java. Further, the problems in Quixbugs are generic functions, such as bucket sort. This makes them easy to solve and hard to decontaminate training data for. Our benchmark, \evalf{}, contains 164 buggy functions for six languages with solutions and unit tests. Further, our coding problems, derived from HumanEval, are very specific, such as keeping track of a bank account balance (see \autoref{fig:operatormisuse}). 

Prior work on evaluating code explanations~\citep{lu2021codexglue,cui2022codeexp} has relied on metrics such as METEOR~\citep{banerjee2005meteor} or BLEU~\citep{papineni2002bleu}. By chaining code explanation with code synthesis, we can evaluate this task using the execution-based pass@$k$ metric overcoming the major limitations of BLEU and other heuristics-based metrics~\citep{reiter2018structured}.

Large-scale benchmarking has proven useful in many areas of natural language processing~\citep{wang2019superglue,kiela2021hateful,bigbench,muennighoff2022mteb}. By producing 18 scores (6 languages across 3 tasks) for 9 models, we take a step towards large-scale benchmarking of code models. However, we lack many models capable of generating code~\citep{black2021gpt,fried2022incoder,black2022gpt,wang2021gpt,biderman2023pythia}. Future work may consider more models or extending \eval{} to new languages or tasks, such as code efficiency~\citep{madaan2023learning,yetistiren2022assessing} or code classification~\citep{khan2023xcodeeval}.

\vspace{-1mm}
\section{Conclusion}

This work studies training and evaluation of Code LLMs that follow instructions. We introduce \data{}, a 4TB dataset of Git commits covering 350 programming languages. We filter this large-scale dataset to create \dataft{}, 2GB of high-quality code with commit messages that assimilate instructions. To enable a comprehensive evaluation of instruction code models, we construct \eval{}, a human-written benchmark covering 3 different tasks for 6 programming languages. We ablate several instruction datasets and find that \dataft{} combined with natural language data leads to the best performance. While our models, \model{} and \modelx{}, are the best permissively licensed Code LLMs available, they are outperformed by closed-source models such as GPT-4. In addition to improving the instruction tuning paradigm, future work should consider training more capable base models.

\newpage

\section*{Acknowledgements}

We thank Hugging Face for providing compute instances. We are extremely grateful to Rodrigo Garcia for the Rust translations, Dimitry Ageev and Calum Bird for help with GPT-4 evaluation, Loubna Ben Allal for help on evaluation, Arjun Guha for insightful discussions on chaining evaluation tasks to avoid evaluating with BLEU, Lewis Tunstall for help on the OASST data, Victor Sanh and Nadav Timor for discussions, Jiaxi Yang for logo editing and domain classification prompting design, \cite{ghosal2023flacuna,zeng2023matters} for design inspiration, Harm de Vries for feedback and all members of BigCode for general support. Finally, we thank every programmer who takes the time to write informative commit messages.


\bibliography{octopack}
\bibliographystyle{iclr2023_conference}

\newpage
\appendix

\part{}
\section*{\centering \LARGE{\textbf{Appendix}}}
\mtcsettitle{parttoc}{Contents}
\parttoc

\newpage


\section{Contributions}
\label{sec:contributions}

Niklas Muennighoff created \data{} and \eval{}, wrote most of the paper and led the project. Qian Liu devised many quality filters, ran SantaCoder ablations, investigated early training decisions and helped edit the paper. Armel Zebaze created the Self-Instruct data and ran numerous ablations. Niklas Muennighoff, Armel Zebaze and Qinkai Zheng created and evaluated \model{} and \modelx{}. Binyuan Hui pretrained SantaCoder, made major contributions to the presentation and helped edit the paper. Terry Yue Zhuo ran GPT-4 evaluations and helped edit the paper. Xiangru Tang provided help on several experiments for evaluation and helped edit the paper. Leandro von Werra provided early guidance, suggested many quality filters and added the commit data to StarCoder pretraining. Niklas Muennighoff, Qian Liu, Binyuan Hui, Swayam Singh and Shayne Longpre conducted the data analysis. Shayne Longpre advised the project and made large contributions to the paper.


\section{Artifacts}
\label{sec:artifacts}

\begin{table*}[htbp]
    \centering
    \resizebox{\textwidth}{!}{
    \begin{tabular}{l|c}
    \toprule
    Model & Public Link\\
    \midrule
    \multicolumn{2}{c}{\textit{Other models}} \\
    \midrule
    Diff Codegen 2B \citep{bradley2023diffmodels} & \url{https://hf.co/CarperAI/diff-codegen-2b-v2} \\
    InstructCodeT5+ \citep{wang2023codet5+} & \url{https://hf.co/Salesforce/instructcodet5p-16b} \\    
    BLOOMZ \citep{muennighoff2022crosslingual} & \url{https://hf.co/bigscience/bloomz} \\    
    StarChat-$\beta$ \citep{Tunstall2023starchat-alpha} & \url{https://hf.co/HuggingFaceH4/starchat-beta} \\    
    CodeGeeX2 \citep{zheng2023codegeex} & \url{https://github.com/THUDM/CodeGeeX2}\\
    SantaCoder \citep{allal2023santacoder} & \url{https://hf.co/bigcode/santacoder} \\ 
    StarCoder \citep{li2023starcoder} & \url{https://hf.co/bigcode/starcoder} \\
    WizardCoder \citep{luo2023wizardcoder} & \url{https://hf.co/WizardLM/WizardCoder-15B-V1.0} \\
    GPT-4 \citep{openai2023gpt4} & \url{https://openai.com/gpt-4} \\    
    \midrule
    \multicolumn{2}{c}{\textit{Data Ablations (\autoref{sec:ablations}) - Data}} \\
    \midrule
    Filtered xP3x code & \url{https://hf.co/datasets/bigcode/xp3x-octopack} \\
    StarCoder Self-Instruct & \url{https://hf.co/datasets/codeparrot/self-instruct-starcoder} \\
    Filtered OASST & \url{https://hf.co/datasets/bigcode/oasst-octopack} \\  
    Manual selection (\autoref{sec:ablations}) & \url{https://hf.co/datasets/bigcode/co-manual} \\
    \midrule
    \multicolumn{2}{c}{\textit{Data Ablations (\autoref{sec:ablations}) - Models}} \\    
    \midrule
    Self-Instruct (SI) & \url{https://hf.co/bigcode/starcoder-s} \\
    OASST (O) & \url{https://hf.co/bigcode/starcoder-o} \\
    SI + O & \url{https://hf.co/bigcode/starcoder-so} \\
    xP3x + O & \url{https://hf.co/bigcode/starcoder-xo} \\
    \dataft{} + O (Formatting) & \url{https://hf.co/bigcode/starcoder-co-format} \\
    \dataft{} + O (Target loss) & \url{https://hf.co/bigcode/starcoder-co-target} \\
    \dataft{} + O (Manual) & \url{https://hf.co/bigcode/starcoder-co-manual} \\
    \dataft{} + xP3x + O & \url{https://hf.co/bigcode/starcoder-cxo} \\
    \dataft{} + xP3x + SI + O & \url{https://hf.co/bigcode/starcoder-cxso} \\
    \midrule
    \multicolumn{2}{c}{\textit{SantaCoder ablations  (\autoref{sec:pretraining}, \autoref{sec:diffformat})}} \\
    \midrule
    Commit format Pretraining & \url{https://hf.co/bigcode/santacoderpack} \\
    Commit format Finetuning & \url{https://hf.co/bigcode/santacoder-cf} \\
    Line diff format Finetuning & \url{https://hf.co/bigcode/santacoder-ldf} \\
    \midrule
    \multicolumn{2}{c}{\textit{Other datasets}} \\
    \midrule
    \data{} Metadata & \url{https://hf.co/datasets/bigcode/commitpackmeta} \\    
    \midrule
    \multicolumn{2}{c}{\textit{Main artifacts}} \\    
    \midrule
    \data{} & \url{https://hf.co/datasets/bigcode/commitpack} \\
    \dataft{} & \url{https://hf.co/datasets/bigcode/commitpackft} \\
    \eval{} & \url{https://hf.co/datasets/bigcode/humanevalpack} \\
    \modelx{} & \url{https://hf.co/bigcode/octogeex} \\
    \model{} & \url{https://hf.co/bigcode/octocoder} \\
    \bottomrule
    \end{tabular}
    }
    \caption{
        \textbf{Used and produced artifacts.}
    }
    \label{tab:artifacts}
\end{table*}

\FloatBarrier
\newpage

\section{\data{} and \dataft{} Languages}
\label{sec:dataapp}

\begin{longtable}{l|ccc|ccc}
    \toprule
 & \multicolumn{3}{c|}{\data{}} & \multicolumn{3}{c}{\dataft{}} \\
Language ($\downarrow$) & MB & Samples & \% (MB) & MB & Samples & \% (MB) \\
\midrule
Total & 3709175.78 & 57700105 & 100.0 & 1545.02 & 702062 & 100.0 \\
\midrule
json & 583293.82 & 3495038 & 15.73 & 86.74 & 39777 & 5.61 \\
xml & 279208.68 & 1923159 & 7.53 & 23.68 & 9337 & 1.53 \\
text & 270662.6 & 1389525 & 7.3 & 66.66 & 46588 & 4.31 \\
javascript & 262824.84 & 5401937 & 7.09 & 125.01 & 52989 & 8.09 \\
objective-c++ & 239009.3 & 32227 & 6.44 & 0.38 & 86 & 0.02 \\
python & 234311.56 & 6189601 & 6.32 & 132.68 & 56025 & 8.59 \\
c & 200876.8 & 2779478 & 5.42 & 21.08 & 8506 & 1.36 \\
c++ & 186585.26 & 2402294 & 5.03 & 14.14 & 4992 & 0.92 \\
markdown & 171849.95 & 7645354 & 4.63 & 131.15 & 62518 & 8.49 \\
java & 127103.45 & 3744377 & 3.43 & 56.28 & 20635 & 3.64 \\
html & 105305.28 & 2366841 & 2.84 & 48.42 & 20214 & 3.13 \\
yaml & 100466.64 & 2592787 & 2.71 & 190.88 & 114320 & 12.35 \\
go & 86444.62 & 1183612 & 2.33 & 12.13 & 5004 & 0.79 \\
csv & 82946.19 & 79268 & 2.24 & 0.53 & 375 & 0.03 \\
php & 74961.64 & 2555419 & 2.02 & 60.22 & 24791 & 3.9 \\
jupyter-notebook & 66854.08 & 94000 & 1.8 & 0.1 & 48 & 0.01 \\
gettext-catalog & 62296.88 & 168327 & 1.68 & 0.13 & 72 & 0.01 \\
sql & 56802.76 & 132772 & 1.53 & 3.74 & 2069 & 0.24 \\
unity3d-asset & 39535.01 & 17867 & 1.07 & 0.16 & 101 & 0.01 \\
typescript & 39254.8 & 572136 & 1.06 & 14.28 & 5868 & 0.92 \\
owl & 36435.46 & 7458 & 0.98 & 0 & 0 & 0.0 \\
ruby & 35830.74 & 2928702 & 0.97 & 195.29 & 69413 & 12.64 \\
c\# & 33669.65 & 923157 & 0.91 & 26.84 & 9346 & 1.74 \\
nix & 33547.92 & 221281 & 0.9 & 3.84 & 1593 & 0.25 \\
shell & 25109.95 & 1017977 & 0.68 & 66.86 & 31217 & 4.33 \\
perl & 21148.93 & 374266 & 0.57 & 4.99 & 2288 & 0.32 \\
tex & 17471.11 & 89283 & 0.47 & 0.56 & 307 & 0.04 \\
css & 16306.63 & 548818 & 0.44 & 9.36 & 5049 & 0.61 \\
restructuredtext & 15613.89 & 494037 & 0.42 & 15.73 & 6560 & 1.02 \\
rust & 15011.3 & 296214 & 0.4 & 7.24 & 2996 & 0.47 \\
groff & 12020.19 & 32923 & 0.32 & 0.4 & 192 & 0.03 \\
ini & 8375.16 & 297100 & 0.23 & 21.04 & 11360 & 1.36 \\
scala & 8325.96 & 316064 & 0.22 & 11.18 & 5040 & 0.72 \\
coffeescript & 6795.14 & 292446 & 0.18 & 16.96 & 5513 & 1.1 \\
haskell & 6306.12 & 217325 & 0.17 & 3.31 & 1389 & 0.21 \\
swift & 5902.72 & 319289 & 0.16 & 16.27 & 4849 & 1.05 \\
lua & 5763.12 & 139091 & 0.16 & 1.85 & 920 & 0.12 \\
svg & 5645.44 & 27095 & 0.15 & 0.25 & 169 & 0.02 \\
gas & 5585.38 & 15121 & 0.15 & 0.34 & 193 & 0.02 \\
ocaml & 5355.4 & 81360 & 0.14 & 0.7 & 333 & 0.05 \\
erlang & 5043.32 & 93685 & 0.14 & 1.19 & 480 & 0.08 \\
makefile & 4238.51 & 343379 & 0.11 & 2.53 & 960 & 0.16 \\
asciidoc & 4138.59 & 96671 & 0.11 & 1.86 & 523 & 0.12 \\
emacs-lisp & 3988.65 & 83228 & 0.11 & 1.97 & 1015 & 0.13 \\
scss & 3944.94 & 288190 & 0.11 & 13.21 & 6829 & 0.86 \\
clojure & 3523.41 & 158674 & 0.09 & 5.07 & 2403 & 0.33 \\
org & 3126.22 & 30198 & 0.08 & 0.27 & 136 & 0.02 \\
common-lisp & 2954.9 & 74628 & 0.08 & 1.45 & 778 & 0.09 \\
diff & 2586.05 & 21021 & 0.07 & 1.48 & 680 & 0.1 \\
groovy & 2569.14 & 110057 & 0.07 & 4.17 & 1486 & 0.27 \\
html+erb & 2450.68 & 225379 & 0.07 & 23.1 & 10910 & 1.5 \\
nesc & 2439.56 & 473 & 0.07 & 0.02 & 7 & 0.0 \\
dart & 2395.8 & 56873 & 0.06 & 1.96 & 765 & 0.13 \\
powershell & 2289.28 & 55381 & 0.06 & 2.06 & 991 & 0.13 \\
f\# & 2289.24 & 66840 & 0.06 & 0.66 & 254 & 0.04 \\
dm & 2223.14 & 55584 & 0.06 & 0.15 & 16 & 0.01 \\
kotlin & 2219.25 & 124266 & 0.06 & 5.37 & 2214 & 0.35 \\
pascal & 2194.68 & 42511 & 0.06 & 0.05 & 25 & 0.0 \\
jsx & 2124.74 & 139148 & 0.06 & 5.5 & 2199 & 0.36 \\
viml & 1948.21 & 74062 & 0.05 & 1.96 & 1063 & 0.13 \\
actionscript & 1844.15 & 28819 & 0.05 & 0.12 & 49 & 0.01 \\
cython & 1736.59 & 25927 & 0.05 & 0.31 & 123 & 0.02 \\
turtle & 1698.95 & 3882 & 0.05 & 0.05 & 21 & 0.0 \\
less & 1616.56 & 88634 & 0.04 & 3.72 & 1360 & 0.24 \\
mathematica & 1475.04 & 925 & 0.04 & 0.01 & 1 & 0.0 \\
xslt & 1441.46 & 27956 & 0.04 & 0.26 & 99 & 0.02 \\
scheme & 1249.24 & 30546 & 0.03 & 0.42 & 213 & 0.03 \\
perl6 & 1223.16 & 12167 & 0.03 & 0.27 & 122 & 0.02 \\
edn & 1186.94 & 2289 & 0.03 & 0.09 & 48 & 0.01 \\
fortran & 1178.55 & 13463 & 0.03 & 0.14 & 70 & 0.01 \\
java-server-pages & 1173.07 & 53574 & 0.03 & 0.45 & 173 & 0.03 \\
standard-ml & 1133.48 & 20097 & 0.03 & 0.15 & 72 & 0.01 \\
cmake & 1132.07 & 58446 & 0.03 & 2.27 & 981 & 0.15 \\
json5 & 1108.2 & 1827 & 0.03 & 0.08 & 33 & 0.01 \\
vala & 1104.51 & 14822 & 0.03 & 0.12 & 50 & 0.01 \\
vue & 1093.8 & 68967 & 0.03 & 1.38 & 587 & 0.09 \\
freemarker & 1032.33 & 36216 & 0.03 & 1.03 & 510 & 0.07 \\
graphql & 1004.84 & 2009 & 0.03 & 0.03 & 17 & 0.0 \\
twig & 958.96 & 39588 & 0.03 & 3.96 & 1610 & 0.26 \\
tcl & 869.83 & 16407 & 0.02 & 0.29 & 103 & 0.02 \\
pod & 859.02 & 14922 & 0.02 & 0.15 & 54 & 0.01 \\
dockerfile & 849.73 & 259379 & 0.02 & 0.1 & 39 & 0.01 \\
yacc & 845.7 & 8230 & 0.02 & 0.01 & 3 & 0.0 \\
postscript & 800.73 & 903 & 0.02 & 0.02 & 9 & 0.0 \\
racket & 796.64 & 16615 & 0.02 & 0.2 & 117 & 0.01 \\
eagle & 785.68 & 2237 & 0.02 & 0.01 & 4 & 0.0 \\
haxe & 772.9 & 28447 & 0.02 & 0.34 & 174 & 0.02 \\
julia & 752.07 & 22695 & 0.02 & 0.31 & 180 & 0.02 \\
handlebars & 740.82 & 49842 & 0.02 & 3.29 & 1429 & 0.21 \\
smarty & 720.94 & 41065 & 0.02 & 1.59 & 737 & 0.1 \\
visual-basic & 681.52 & 10511 & 0.02 & 0.15 & 48 & 0.01 \\
literate-haskell & 673.74 & 10729 & 0.02 & 0.02 & 7 & 0.0 \\
smalltalk & 665.89 & 11741 & 0.02 & 0.46 & 284 & 0.03 \\
isabelle & 655.82 & 8359 & 0.02 & 0.01 & 2 & 0.0 \\
nimrod & 652.86 & 12023 & 0.02 & 0.24 & 67 & 0.02 \\
zig & 621.38 & 4290 & 0.02 & 0.01 & 4 & 0.0 \\
m4 & 603.58 & 12465 & 0.02 & 0.26 & 101 & 0.02 \\
max & 603.56 & 2259 & 0.02 & 0 & 0 & 0.0 \\
elixir & 558.12 & 35473 & 0.02 & 2.35 & 1150 & 0.15 \\
mako & 543.01 & 8943 & 0.01 & 0.76 & 170 & 0.05 \\
arduino & 534.18 & 32350 & 0.01 & 0.46 & 225 & 0.03 \\
jade & 531.4 & 46993 & 0.01 & 2.35 & 1119 & 0.15 \\
haml & 502.01 & 74792 & 0.01 & 10.74 & 4415 & 0.7 \\
elm & 481.97 & 18542 & 0.01 & 0.62 & 265 & 0.04 \\
purebasic & 474.28 & 36 & 0.01 & 0.02 & 5 & 0.0 \\
coldfusion & 470.78 & 9263 & 0.01 & 0.02 & 9 & 0.0 \\
lean & 470.03 & 7507 & 0.01 & 0.02 & 3 & 0.0 \\
r & 454.32 & 12858 & 0.01 & 0.23 & 121 & 0.01 \\
cuda & 437.67 & 11450 & 0.01 & 0.07 & 25 & 0.0 \\
textile & 425.12 & 18491 & 0.01 & 0.18 & 61 & 0.01 \\
robotframework & 421.61 & 9211 & 0.01 & 0.21 & 85 & 0.01 \\
abap & 409.62 & 1955 & 0.01 & 0.01 & 1 & 0.0 \\
rdoc & 397.03 & 38760 & 0.01 & 0.55 & 270 & 0.04 \\
llvm & 382.2 & 10727 & 0.01 & 1.6 & 780 & 0.1 \\
ada & 380.7 & 13258 & 0.01 & 0.73 & 265 & 0.05 \\
batchfile & 372.16 & 43674 & 0.01 & 2.98 & 1466 & 0.19 \\
qml & 361.45 & 19360 & 0.01 & 0.94 & 368 & 0.06 \\
jasmin & 359.82 & 4782 & 0.01 & 0.05 & 9 & 0.0 \\
assembly & 343.62 & 8126 & 0.01 & 0.17 & 105 & 0.01 \\
g-code & 334.96 & 3690 & 0.01 & 0.04 & 7 & 0.0 \\
cucumber & 331.38 & 26677 & 0.01 & 2.59 & 976 & 0.17 \\
html+php & 323.35 & 18381 & 0.01 & 0.33 & 150 & 0.02 \\
kicad & 321.94 & 759 & 0.01 & 0 & 0 & 0.0 \\
api-blueprint & 317.85 & 4765 & 0.01 & 0.06 & 23 & 0.0 \\
eiffel & 311.48 & 373 & 0.01 & 0.01 & 2 & 0.0 \\
toml & 292.68 & 63517 & 0.01 & 5.58 & 3424 & 0.36 \\
modelica & 284.62 & 2611 & 0.01 & 0.04 & 15 & 0.0 \\
bitbake & 277.58 & 43239 & 0.01 & 4.46 & 1308 & 0.29 \\
lex & 275.96 & 705 & 0.01 & 0 & 0 & 0.0 \\
stylus & 273.06 & 21967 & 0.01 & 0.95 & 480 & 0.06 \\
protocol-buffer & 254.12 & 9202 & 0.01 & 0.52 & 181 & 0.03 \\
unknown & 252.23 & 30570 & 0.01 & 3.05 & 1597 & 0.2 \\
nit & 244.54 & 4951 & 0.01 & 0.02 & 3 & 0.0 \\
factor & 241.19 & 15378 & 0.01 & 0.36 & 113 & 0.02 \\
xs & 239.04 & 3215 & 0.01 & 0.02 & 7 & 0.0 \\
sass & 230.65 & 23144 & 0.01 & 1.36 & 705 & 0.09 \\
pir & 230.2 & 6231 & 0.01 & 0.08 & 23 & 0.01 \\
html+django & 217.04 & 10535 & 0.01 & 0.85 & 399 & 0.06 \\
mediawiki & 214.32 & 10188 & 0.01 & 0.08 & 33 & 0.01 \\
logos & 212.3 & 1733 & 0.01 & 0.04 & 19 & 0.0 \\
genshi & 209.3 & 956 & 0.01 & 0.02 & 3 & 0.0 \\
coldfusion-cfc & 208.16 & 4410 & 0.01 & 0.05 & 20 & 0.0 \\
xtend & 179.54 & 7775 & 0.0 & 0.13 & 55 & 0.01 \\
sqf & 168.66 & 7778 & 0.0 & 0.09 & 45 & 0.01 \\
vhdl & 155.95 & 2185 & 0.0 & 0.02 & 5 & 0.0 \\
antlr & 143.55 & 3651 & 0.0 & 0.03 & 15 & 0.0 \\
systemverilog & 140.19 & 3944 & 0.0 & 0.08 & 35 & 0.01 \\
hcl & 136.75 & 13379 & 0.0 & 0.91 & 421 & 0.06 \\
asp & 136.1 & 4286 & 0.0 & 0.09 & 22 & 0.01 \\
nsis & 129.12 & 4048 & 0.0 & 0.06 & 15 & 0.0 \\
inform-7 & 120.19 & 184 & 0.0 & 0.01 & 2 & 0.0 \\
slim & 119.04 & 18726 & 0.0 & 2.06 & 1052 & 0.13 \\
groovy-server-pages & 117.37 & 6695 & 0.0 & 0.07 & 25 & 0.0 \\
ceylon & 116.14 & 7256 & 0.0 & 0.1 & 49 & 0.01 \\
fish & 111.28 & 15351 & 0.0 & 1.33 & 813 & 0.09 \\
processing & 108.58 & 5912 & 0.0 & 0.07 & 35 & 0.0 \\
component-pascal & 105.5 & 43 & 0.0 & 0 & 0 & 0.0 \\
lasso & 104.17 & 67 & 0.0 & 0 & 0 & 0.0 \\
glsl & 99.49 & 9478 & 0.0 & 0.34 & 164 & 0.02 \\
saltstack & 98.2 & 12314 & 0.0 & 1.41 & 617 & 0.09 \\
xbase & 94.42 & 1670 & 0.0 & 0.01 & 3 & 0.0 \\
autohotkey & 94.22 & 1452 & 0.0 & 0.02 & 15 & 0.0 \\
liquid & 93.79 & 2651 & 0.0 & 0.09 & 30 & 0.01 \\
purescript & 92.41 & 5024 & 0.0 & 0.17 & 80 & 0.01 \\
agda & 92.06 & 4956 & 0.0 & 0.02 & 10 & 0.0 \\
inno-setup & 91.36 & 3014 & 0.0 & 0.06 & 16 & 0.0 \\
oz & 90.48 & 1551 & 0.0 & 0.03 & 8 & 0.0 \\
chapel & 89.62 & 26447 & 0.0 & 0.04 & 20 & 0.0 \\
arc & 87.21 & 758 & 0.0 & 0.01 & 2 & 0.0 \\
opencl & 86.43 & 2489 & 0.0 & 0.05 & 23 & 0.0 \\
graphviz-dot & 85.8 & 1525 & 0.0 & 0.07 & 35 & 0.0 \\
pawn & 85.42 & 580 & 0.0 & 0.01 & 3 & 0.0 \\
jsoniq & 75.15 & 1343 & 0.0 & 0.01 & 6 & 0.0 \\
bluespec & 72.38 & 2500 & 0.0 & 0.01 & 2 & 0.0 \\
smali & 71.38 & 174 & 0.0 & 0 & 0 & 0.0 \\
krl & 69.87 & 1879 & 0.0 & 0.02 & 4 & 0.0 \\
maple & 68.28 & 1311 & 0.0 & 0.01 & 2 & 0.0 \\
unrealscript & 67.67 & 585 & 0.0 & 0.01 & 1 & 0.0 \\
ooc & 63.19 & 3416 & 0.0 & 0.04 & 15 & 0.0 \\
pure-data & 62.62 & 603 & 0.0 & 0.01 & 1 & 0.0 \\
xquery & 61.96 & 2237 & 0.0 & 0.08 & 39 & 0.01 \\
dcl & 59.64 & 833 & 0.0 & 0.04 & 19 & 0.0 \\
moonscript & 59.21 & 1951 & 0.0 & 0.02 & 10 & 0.0 \\
awk & 57.18 & 2206 & 0.0 & 0.1 & 52 & 0.01 \\
pike & 52.87 & 1262 & 0.0 & 0.02 & 6 & 0.0 \\
livescript & 51.23 & 5194 & 0.0 & 0.13 & 63 & 0.01 \\
solidity & 50.86 & 3689 & 0.0 & 0.08 & 37 & 0.01 \\
monkey & 48.26 & 1367 & 0.0 & 0.02 & 4 & 0.0 \\
jsonld & 48.01 & 462 & 0.0 & 0.02 & 6 & 0.0 \\
zephir & 42.68 & 1265 & 0.0 & 0.02 & 4 & 0.0 \\
crystal & 41.92 & 4217 & 0.0 & 0.35 & 182 & 0.02 \\
rhtml & 41.02 & 4551 & 0.0 & 0.35 & 135 & 0.02 \\
stata & 40.68 & 1344 & 0.0 & 0.02 & 10 & 0.0 \\
idris & 39.9 & 3025 & 0.0 & 0.13 & 38 & 0.01 \\
raml & 39.39 & 948 & 0.0 & 0.03 & 9 & 0.0 \\
openscad & 37.73 & 2178 & 0.0 & 0.05 & 21 & 0.0 \\
red & 35.26 & 1108 & 0.0 & 0.01 & 1 & 0.0 \\
c2hs-haskell & 34.47 & 1021 & 0.0 & 0.01 & 2 & 0.0 \\
cycript & 33.96 & 197 & 0.0 & 0 & 0 & 0.0 \\
applescript & 33.51 & 1304 & 0.0 & 0.04 & 19 & 0.0 \\
mupad & 32.49 & 178 & 0.0 & 0.02 & 4 & 0.0 \\
literate-agda & 31.38 & 567 & 0.0 & 0.01 & 1 & 0.0 \\
boo & 31.17 & 26289 & 0.0 & 0.01 & 2 & 0.0 \\
sourcepawn & 29.53 & 717 & 0.0 & 0.01 & 3 & 0.0 \\
qmake & 29.51 & 3632 & 0.0 & 0.32 & 140 & 0.02 \\
ragel-in-ruby-host & 28.3 & 888 & 0.0 & 0.01 & 4 & 0.0 \\
io & 27.95 & 1247 & 0.0 & 0.01 & 4 & 0.0 \\
desktop & 27.65 & 5021 & 0.0 & 0.36 & 186 & 0.02 \\
propeller-spin & 26.77 & 625 & 0.0 & 0.01 & 1 & 0.0 \\
thrift & 26.75 & 1007 & 0.0 & 0.08 & 28 & 0.01 \\
volt & 25.05 & 1660 & 0.0 & 0.02 & 9 & 0.0 \\
xproc & 24.21 & 914 & 0.0 & 0.02 & 3 & 0.0 \\
igor-pro & 23.75 & 388 & 0.0 & 0.01 & 1 & 0.0 \\
lolcode & 23.74 & 24861 & 0.0 & 0 & 0 & 0.0 \\
html+eex & 21.41 & 2100 & 0.0 & 0.29 & 135 & 0.02 \\
logtalk & 20.43 & 1035 & 0.0 & 0.06 & 21 & 0.0 \\
mirah & 20.1 & 706 & 0.0 & 0.04 & 16 & 0.0 \\
gnuplot & 19.68 & 889 & 0.0 & 0.03 & 17 & 0.0 \\
literate-coffeescript & 19.02 & 1041 & 0.0 & 0.05 & 19 & 0.0 \\
jflex & 18.61 & 555 & 0.0 & 0.01 & 1 & 0.0 \\
emberscript & 18.39 & 1024 & 0.0 & 0.02 & 7 & 0.0 \\
cobol & 17.0 & 24953 & 0.0 & 0 & 0 & 0.0 \\
yang & 16.94 & 597 & 0.0 & 0.02 & 6 & 0.0 \\
rebol & 16.47 & 239 & 0.0 & 0.01 & 3 & 0.0 \\
linker-script & 16.08 & 1604 & 0.0 & 0.08 & 37 & 0.01 \\
cartocss & 15.92 & 555 & 0.0 & 0.01 & 3 & 0.0 \\
urweb & 13.07 & 304 & 0.0 & 0.02 & 6 & 0.0 \\
rmarkdown & 13.03 & 750 & 0.0 & 0 & 0 & 0.0 \\
darcs-patch & 13.01 & 80 & 0.0 & 0 & 0 & 0.0 \\
csound & 12.85 & 229 & 0.0 & 0.01 & 4 & 0.0 \\
squirrel & 12.84 & 531 & 0.0 & 0.01 & 4 & 0.0 \\
apl & 12.56 & 586 & 0.0 & 0.02 & 7 & 0.0 \\
hlsl & 12.17 & 1529 & 0.0 & 0.03 & 11 & 0.0 \\
latte & 11.89 & 1380 & 0.0 & 0.02 & 7 & 0.0 \\
pony & 11.84 & 624 & 0.0 & 0.05 & 16 & 0.0 \\
ioke & 10.86 & 373 & 0.0 & 0.04 & 25 & 0.0 \\
hy & 10.51 & 879 & 0.0 & 0.04 & 12 & 0.0 \\
uno & 10.36 & 628 & 0.0 & 0.01 & 2 & 0.0 \\
pan & 10.34 & 637 & 0.0 & 0.05 & 23 & 0.0 \\
xojo & 10.31 & 642 & 0.0 & 0 & 0 & 0.0 \\
papyrus & 10.26 & 130 & 0.0 & 0 & 0 & 0.0 \\
stan & 10.25 & 540 & 0.0 & 0 & 0 & 0.0 \\
slash & 9.9 & 640 & 0.0 & 0.01 & 4 & 0.0 \\
supercollider & 9.8 & 318 & 0.0 & 0.01 & 2 & 0.0 \\
vcl & 9.46 & 747 & 0.0 & 0.04 & 18 & 0.0 \\
smt & 9.03 & 117 & 0.0 & 0.01 & 3 & 0.0 \\
glyph & 8.95 & 7 & 0.0 & 0 & 0 & 0.0 \\
wisp & 8.74 & 262 & 0.0 & 0.01 & 3 & 0.0 \\
renpy & 8.3 & 421 & 0.0 & 0.02 & 3 & 0.0 \\
clips & 7.73 & 450 & 0.0 & 0 & 0 & 0.0 \\
dns-zone & 7.56 & 54 & 0.0 & 0.01 & 2 & 0.0 \\
sas & 7.54 & 269 & 0.0 & 0.01 & 1 & 0.0 \\
rouge & 7.2 & 396 & 0.0 & 0.1 & 41 & 0.01 \\
ec & 7.03 & 94 & 0.0 & 0 & 0 & 0.0 \\
dylan & 6.82 & 280 & 0.0 & 0.01 & 2 & 0.0 \\
tcsh & 6.52 & 748 & 0.0 & 0.02 & 10 & 0.0 \\
aspectj & 6.33 & 451 & 0.0 & 0.02 & 8 & 0.0 \\
netlogo & 6.3 & 140 & 0.0 & 0 & 0 & 0.0 \\
gap & 6.1 & 46 & 0.0 & 0 & 0 & 0.0 \\
fancy & 5.95 & 675 & 0.0 & 0.02 & 8 & 0.0 \\
coq & 5.74 & 80 & 0.0 & 0 & 0 & 0.0 \\
click & 5.74 & 9 & 0.0 & 0 & 0 & 0.0 \\
capn-proto & 5.64 & 330 & 0.0 & 0.04 & 12 & 0.0 \\
flux & 5.57 & 47 & 0.0 & 0.01 & 3 & 0.0 \\
forth & 5.51 & 265 & 0.0 & 0.01 & 2 & 0.0 \\
ats & 5.42 & 383 & 0.0 & 0.01 & 3 & 0.0 \\
netlinx & 5.17 & 144 & 0.0 & 0.01 & 1 & 0.0 \\
clean & 5.07 & 171 & 0.0 & 0.01 & 1 & 0.0 \\
parrot-assembly & 4.66 & 227 & 0.0 & 0.01 & 2 & 0.0 \\
alloy & 4.64 & 203 & 0.0 & 0 & 0 & 0.0 \\
lfe & 4.58 & 287 & 0.0 & 0.02 & 6 & 0.0 \\
gdscript & 4.49 & 460 & 0.0 & 0.03 & 9 & 0.0 \\
augeas & 4.44 & 395 & 0.0 & 0.04 & 13 & 0.0 \\
sparql & 4.4 & 1036 & 0.0 & 0.04 & 23 & 0.0 \\
lilypond & 4.31 & 265 & 0.0 & 0.01 & 6 & 0.0 \\
scilab & 4.09 & 375 & 0.0 & 0.02 & 10 & 0.0 \\
autoit & 4.06 & 279 & 0.0 & 0 & 0 & 0.0 \\
myghty & 3.86 & 105 & 0.0 & 0 & 0 & 0.0 \\
blitzmax & 3.74 & 220 & 0.0 & 0.01 & 1 & 0.0 \\
creole & 3.42 & 337 & 0.0 & 0.01 & 2 & 0.0 \\
harbour & 3.34 & 107 & 0.0 & 0.01 & 1 & 0.0 \\
piglatin & 3.17 & 513 & 0.0 & 0.02 & 11 & 0.0 \\
opa & 3.16 & 211 & 0.0 & 0 & 0 & 0.0 \\
sage & 3.03 & 414 & 0.0 & 0.01 & 1 & 0.0 \\
ston & 2.85 & 414 & 0.0 & 0.01 & 6 & 0.0 \\
maxscript & 2.8 & 47 & 0.0 & 0 & 0 & 0.0 \\
lsl & 2.68 & 74 & 0.0 & 0.01 & 3 & 0.0 \\
gentoo-ebuild & 2.58 & 601 & 0.0 & 0.06 & 16 & 0.0 \\
nu & 2.38 & 170 & 0.0 & 0.01 & 2 & 0.0 \\
bro & 2.34 & 333 & 0.0 & 0.01 & 3 & 0.0 \\
xc & 2.02 & 88 & 0.0 & 0 & 0 & 0.0 \\
j & 1.81 & 142 & 0.0 & 0 & 0 & 0.0 \\
metal & 1.72 & 151 & 0.0 & 0.02 & 4 & 0.0 \\
mms & 1.54 & 91 & 0.0 & 0.01 & 1 & 0.0 \\
webidl & 1.51 & 96 & 0.0 & 0.05 & 6 & 0.0 \\
tea & 1.47 & 29 & 0.0 & 0 & 0 & 0.0 \\
redcode & 1.27 & 149 & 0.0 & 0 & 0 & 0.0 \\
shen & 1.2 & 71 & 0.0 & 0 & 0 & 0.0 \\
pov-ray-sdl & 1.14 & 104 & 0.0 & 0.01 & 5 & 0.0 \\
x10 & 1.01 & 33 & 0.0 & 0 & 0 & 0.0 \\
brainfuck & 0.96 & 167 & 0.0 & 0.01 & 2 & 0.0 \\
ninja & 0.95 & 187 & 0.0 & 0.03 & 14 & 0.0 \\
golo & 0.9 & 115 & 0.0 & 0 & 0 & 0.0 \\
webassembly & 0.86 & 83 & 0.0 & 0 & 0 & 0.0 \\
self & 0.82 & 15 & 0.0 & 0 & 0 & 0.0 \\
labview & 0.81 & 61 & 0.0 & 0 & 0 & 0.0 \\
octave & 0.8 & 12 & 0.0 & 0 & 0 & 0.0 \\
pogoscript & 0.8 & 74 & 0.0 & 0 & 0 & 0.0 \\
d & 0.8 & 20 & 0.0 & 0 & 0 & 0.0 \\
http & 0.74 & 140 & 0.0 & 0.03 & 19 & 0.0 \\
ecl & 0.66 & 48 & 0.0 & 0.01 & 4 & 0.0 \\
chuck & 0.58 & 99 & 0.0 & 0 & 0 & 0.0 \\
gosu & 0.52 & 60 & 0.0 & 0 & 0 & 0.0 \\
parrot & 0.52 & 17 & 0.0 & 0 & 0 & 0.0 \\
opal & 0.47 & 69 & 0.0 & 0 & 0 & 0.0 \\
objective-j & 0.46 & 37 & 0.0 & 0 & 0 & 0.0 \\
kit & 0.41 & 48 & 0.0 & 0 & 0 & 0.0 \\
gams & 0.38 & 18 & 0.0 & 0 & 0 & 0.0 \\
prolog & 0.28 & 35 & 0.0 & 0 & 0 & 0.0 \\
clarion & 0.27 & 13 & 0.0 & 0 & 0 & 0.0 \\
mask & 0.25 & 37 & 0.0 & 0.01 & 4 & 0.0 \\
brightscript & 0.24 & 28 & 0.0 & 0 & 0 & 0.0 \\
scaml & 0.18 & 31 & 0.0 & 0.01 & 1 & 0.0 \\
matlab & 0.16 & 29 & 0.0 & 0 & 0 & 0.0 \\
idl & 0.15 & 1 & 0.0 & 0 & 0 & 0.0 \\
ags-script & 0.12 & 31 & 0.0 & 0 & 0 & 0.0 \\
lookml & 0.12 & 10 & 0.0 & 0 & 0 & 0.0 \\
apacheconf & 0.11 & 59 & 0.0 & 0.01 & 2 & 0.0 \\
oxygene & 0.1 & 9 & 0.0 & 0 & 0 & 0.0 \\
txl & 0.1 & 3 & 0.0 & 0 & 0 & 0.0 \\
gf & 0.09 & 39 & 0.0 & 0 & 0 & 0.0 \\
renderscript & 0.06 & 54 & 0.0 & 0 & 0 & 0.0 \\
mtml & 0.05 & 13 & 0.0 & 0.01 & 2 & 0.0 \\
unified-parallel-c & 0.05 & 6 & 0.0 & 0 & 0 & 0.0 \\
dogescript & 0.04 & 10 & 0.0 & 0 & 0 & 0.0 \\
gentoo-eclass & 0.04 & 6 & 0.0 & 0 & 0 & 0.0 \\
zimpl & 0.04 & 7 & 0.0 & 0 & 0 & 0.0 \\
irc-log & 0.04 & 9 & 0.0 & 0 & 0 & 0.0 \\
fantom & 0.03 & 11 & 0.0 & 0 & 0 & 0.0 \\
numpy & 0.03 & 1 & 0.0 & 0 & 0 & 0.0 \\
cirru & 0.02 & 4 & 0.0 & 0 & 0 & 0.0 \\
xpages & 0.02 & 7 & 0.0 & 0.01 & 1 & 0.0 \\
nginx & 0.02 & 6 & 0.0 & 0.01 & 2 & 0.0 \\
objdump & 0.02 & 1 & 0.0 & 0 & 0 & 0.0 \\
python-traceback & 0.02 & 10 & 0.0 & 0 & 0 & 0.0 \\
realbasic & 0.01 & 1 & 0.0 & 0 & 0 & 0.0 \\
befunge & 0.01 & 2 & 0.0 & 0 & 0 & 0.0 \\
bison & 0.01 & 1 & 0.0 & 0 & 0 & 0.0 \\
m & 0.01 & 1 & 0.0 & 0 & 0 & 0.0 \\
omgrofl & 0.01 & 1 & 0.0 & 0 & 0 & 0.0 \\
    \bottomrule
    \caption{\textbf{Programming language distribution of \data{} and \dataft{}.} Shortcuts: MB=Megabytes, owl=web-ontology-language, pir=parrot-internal-representation, dcl=digital-command-language, mms=module-management-system, gf=grammatical-framework}
    \label{tab:data}
\end{longtable}    


\section{Dataset Creation}
\label{sec:datacreation}

\paragraph{\data} We use the GitHub archive available on GCP which contains metadata from GitHub commits up to 2016.\footnote{https://www.gharchive.org/} It contains around 3TB of GitHub activity data for more than 2.8 million GitHub repositories including more than 145 million unique commits, over 2 billion different file paths and the contents of the latest revision for 163 million files.\footnote{https://github.blog/2016-06-29-making-open-source-data-more-available/} We apply the filters in \autoref{tab:filtera} to this dataset. The resulting dataset containing only metadata is uploaded at \url{https://hf.co/datasets/bigcode/commitpackmeta}. As the activity dataset only contains commit ids without the actual code changes, we scrape the code from GitHub. We use the metadata and the GitHub API to scrape the changed file prior and after the respective commit. Some repositories referenced in the activity data are no longer accessible, thus we discard them. This results in \data{} with approximately 4 terabytes uploaded at \url{https://hf.co/datasets/bigcode/commitpack}.


\begin{table}[htbp]
\centering
\begin{tabular}{p{2cm} p{11cm}}
\toprule
\textbf{Description} & \textbf{Details} \\
\midrule
License & Only keep samples licensed as MIT, Artistic-2.0, ISC, CC0-1.0, EPL-1.0, MPL-2.0, Apache-2.0, BSD-3-Clause, AGPL-3.0, LGPL-2.1, BSD-2-Clause or without license. \\
Length & Only keep code where the commit message has at least 5 and at most 10,000 characters \\
Noise & Remove code where the lowercased commit message is any of 'add files via upload', "can't you see i'm updating the time?", 'commit', 'create readme.md', 'dummy', 'first commit', 'heartbeat update', 'initial commit', 'mirroring from micro.blog.', 'no message', 'pi push', 'readme', 'update', 'updates', 'update \_config.yaml', 'update index.html', 'update readme.md', 'update readme', 'updated readme', 'update log', 'update data.js', 'update data.json', 'update data.js', 'pi push' or starts with 'merge' \\
Single file & Remove samples that contain changes across multiple files\\
Opt-out & Remove samples from repositories owned by users that opted out of The Stack~\citep{kocetkov2022stack}\\
\bottomrule
\end{tabular}
\caption{\textbf{\data{} filters.}}
\label{tab:filtera}
\end{table}


\paragraph{\dataft{}} Prior work has shown the importance of careful data filtering to maintain quality \citep{yin2018mining,dhole2021nl,laurenccon2022bigscience,longpre2023pretrainer,singh2024aya}. To create a smaller version focused on commits that resemble high-quality instructions, we further filter \data{} to create \dataft{} using the steps outlined in \autoref{tab:filterb}. We also checked for any contamination with HumanEval~\citep{chen2021evaluating} but did not find any solution or docstring present in \dataft{}. This is likely because our commit data only goes up to 2016, which is several years prior to the release of HumanEval. Our filters reduce the dataset by a factor of around 1000 resulting in close to 2 gigabytes across 277 languages. To gain a deeper understanding of the rich content within \dataft{}, we analyze commits on its Python subset (56K samples). We first collect the most prevalent commit domain by prompting GPT-4 with: \texttt{"I'd like to know the main types of commits on Github and aim to cover as comprehensively as possible."}. Subsequently, we use GPT-4 to classify each sample using the prompt in \autoref{fig:commitdomain}. The task distribution is visualized in \autoref{fig:data}.

\begin{table}[htbp]
\centering
\begin{tabular}{p{2cm} p{11cm}}
\toprule
\textbf{Description} & \textbf{Details} \\
\midrule
Length & Remove samples where the before code has more than 50,000 characters\\
Length & Remove samples where the after code has 0 characters\\
Difference & Remove samples where the before and after code are the same (e.g. file name changes)\\
Difference & Remove samples that contain a hashtag (to avoid references to issues) \\
Extension & Remove samples where the filename of the code after has an atypical extension for the programming language (e.g. only keep '.py' for Python) \\
Filename & Remove samples where the filename is contained in the commit message (as we do not use the filename in finetuning)\\
Length & Only keep samples where the commit message has more than 10 and less than 1000 characters\\
Words & Only keep samples where the commit message can be split into more than 4 and less than 1000 space-separated words\\
Clean & Remove any appearances of '[skip ci]', '[ci skip]', sequences at the beginning or end that are in brackets, sequences at the beginning that end with ':' and strip whitespace at the beginning or end\\
Capitalized & Only keep samples where the message starts with an uppercase letter\\
Tokens & Only keep samples where the concatenation of the code before, a special token and the code after has at least 50 tokens and at most 768 tokens according to the StarCoder tokenizer\\
Instructions & Only keep samples where the lowercased commit message starts with any of the words in \autoref{tab:filterwords} \\
Noise & Remove samples where the lowercased commit message contains any of 'auto commit', 'update contributing', '<?xml', 'merge branch', 'merge pull request', 'signed-off-by', "fix that bug where things didn't work but now they should", "put the thingie in the thingie", "add a beter commit message", "code review", "//codereview", "work in progress", "wip", "https://", "http://", "| leetcode", "cdpcp", " i ", "i've" , "i'm" or both "thanks to" and "for" \\
Regex & Remove samples where the lowercased commit message has a match for any of the regular expressions \verb/(?:v)?\d+\.\d+\.\d+(?=$|\S)/, \verb/^[a-f0-9]+(?:-[a-f0-9]+)*$/, \verb/([a-f0-9]{40})/, \verb/issue\s*\d+/, \verb/bug\s*\d+/ or \verb/feature\s*\d+/ \\
Downsample & With 90\% probability remove samples where the commit message starts with "Bump", "Set version" or "Update version" \\
\bottomrule
\end{tabular}
\caption{\textbf{\dataft{} filters applied to \data{}.} With the commit message we refer to the commit message subject only, not the body.}
\label{tab:filterb}
\end{table}

\begin{table}[htbp]
\centering
\begin{tabular}{p{13cm}}
\toprule
"abort', 'accelerate', 'access', 'accumulate', 'add', 'address', 'adjust', 'advance', 'align', 'allot', 'allow', 'amplify', 'annotate', 'append', 'apply', 'archive', 'arrange', 'attach', 'augment', 'automate', 'backup', 'boost', 'break', 'bring', 'brush up', 'build', 'bump', 'call', 'change', 'check', 'choose', 'clarify', 'clean', 'clear', 'clone', 'comment', 'complete', 'compress', 'concatenate', 'configure', 'connect', 'consolidate', 'convert', 'copy', 'correct', 'cover', 'create', 'customize', 'cut', 'deal with', 'debug', 'decipher', 'declare', 'decommission', 'decomplexify', 'decompress', 'decrease', 'decrypt', 'define', 'delete', 'deploy', 'designate', 'destroy', 'detach', 'determine', 'develop', 'diminish', 'disable', 'discard', 'disentangle', 'dismantle', 'divide', 'document', 'downgrade', 'drop', 'duplicate', 'edit', 'embed', 'emphasize', 'enable', 'encrypt', 'enforce', 'enhance', 'enlarge', 'enumerate', 'eradicate', 'escalate', 'establish', 'exclude', 'exit', 'expand', 'expedite', 'expire', 'extend', 'facilitate', 'fix', 'format', 'gather', 'generalize', 'halt', 'handle', 'hasten', 'hide', 'implement', 'improve', 'include', 'increase', 'increment', 'indent', 'index', 'inflate', 'initialize', 'insert', 'install', 'integrate', 'interpolate', 'interrupt', 'introduce', 'isolate', 'join', 'kill', 'leverage', 'load', 'magnify', 'maintain', 'make', 'manage', 'mark', 'mask', 'mend', 'merge', 'migrate', 'modify', 'monitor', 'move', 'multiply', 'normalize', 'optimize', 'orchestrate', 'order', 'package', 'paraphrase', 'paste', 'patch', 'plug ', 'prepare', 'prepend', 'print', 'provision', 'purge', 'put', 'quit', 'raise', 'read', 'reannotate', 'rearrange', 'rebase', 'reboot', 'rebuild', 'recomment', 'recompile', 'reconfigure', 'reconnect', 'rectify', 'redact', 'redefine', 'reduce', 'refactor', 'reformat', 'refresh', 'reimplement', 'reinforce', 'relocate', 'remove', 'rename', 'reorder', 'reorganize', 'repackage', 'repair', 'rephrase', 'replace', 'reposition', 'reschedule', 'reset', 'reshape', 'resolve', 'restructure', 'return', 'revert', 'revise', 'revoke', 'reword', 'rework', 'rewrite', 'rollback', 'save', 'scale', 'scrub', 'secure', 'select', 'send', 'set', 'settle', 'simplify', 'solve', 'sort', 'speed up', 'split', 'stabilize', 'standardize', 'stipulate', 'stop', 'store', 'streamline', 'strengthen', 'structure', 'substitute', 'subtract', 'support', 'swap', 'switch', 'synchronize', 'tackle', 'tag', 'terminate', 'test', 'throw', 'tidy', 'transform', 'transpose', 'trim', 'troubleshoot', 'truncate', 'tweak', 'unblock', 'uncover', 'undo', 'unify', 'uninstall', 'unplug', 'unpublish', 'unravel', 'unstage', 'unsync', 'untangle', 'unwind', 'update', 'upgrade', 'use', 'validate', 'verify', 'watch', 'watermark', 'whitelist', 'withdraw', 'work', 'write" \\
\bottomrule
\end{tabular}
\caption{\textbf{Commit message starting words allowed in \dataft{}.}}
\label{tab:filterwords}
\end{table}

\begin{figure}[htbp]
\hrulefill

Please categorize the following commit message, which may fall into more than one category. \\

\#\#\# Category \\
Bug fixes, New features, Refactoring/code cleanup, Documentation, Testing, User interface, Dependencies, Configuration, Build system/tooling, Performance improvements, Formatting/Linting, Security, Technical debt repayment, Release management, Accessibility, Deprecation, Logging/Instrumentation, Internationalization \\

\#\#\# Commit Message \\
Add the blacklist checking to the bulk \\

\#\#\# Classification \\
Bug fixes, New features \\

\#\#\# Commit Message \\
\{COMMIT\_MESSAGE\} 

\#\#\# Classification \\

\hrulefill
\caption[]{\textbf{GPT-4 1-shot prompt for classifying commits in \dataft{}.}}
\label{fig:commitdomain}
\end{figure}

\paragraph{xP3x} We use a subset of xP3x~\citep{muennighoff2022crosslingual} focusing on code datasets consisting of APPS~\citep{hendrycks2021measuring}, CodeContests~\citep{li2022competition}, Jupyter Code Pairs,\footnote{\url{https://hf.co/datasets/codeparrot/github-jupyter-text-code-pairs}} MBPP~\citep{austin2021program}, XLCoST~\citep{zhu2022xlcost}, Code Complex~\citep{JeonBHHK22}, Docstring Corpus \citep{barone2017parallel}, Great Code \citep{hellendoorn2019global} and State Changes.\footnote{\url{https://hf.co/datasets/Fraser/python-state-changes}}

\paragraph{OASST} We reuse a filtered variant of OASST~\citep{kopf2023openassistant} from prior work~\citep{dettmers2023qlora} and apply additional filters to remove responses that refuse to comply with the user request. To compute the programming languages and code fraction for OASST depicted in \autoref{tab:ablationsdata}, we count all responses containing e.g. \verb|```python| or \verb|```py| for the Python programming language. There are code samples that are not enclosed in backticks or do not specify the language, thus we are likely underestimating the actual fraction of code data for OASST in \autoref{tab:ablationsdata}.

\section{Comparing Data Before and After Filtering}
\label{sec:compare}

In~\autoref{tab:filterstats} we compare word statistics prior to and after filtering \data{} to create \dataft{}. The mean commit subject and message length increases suggesting that messages are more informative in \dataft{}. The code lengths decrease significantly as we limit the number of allowed tokens in the filters in \autoref{tab:filterb}. This is intended, as we would like to maximize the amount of training signal per token. The code before and after the commit are usually largely the same. By filtering for short samples, we ensure that there are more differences between the code before and after, thus making the model learn faster. The percentage of code changed between pre- and post-commit is $77.6 / 59.1 = 1.31$ (a 31\% increase) as opposed to $3269.8 / 3269.9 = 1.007$ (a 0.7\% increase). Thus, the filtered data carries significantly more signal per token with fewer repetitions of the code prior to the commit.

\begin{table*}[htbp]
    \centering
    \begin{tabular}{l|ccc}
    \toprule
    Metric & Before Filter & After Filter & Difference \\
    \midrule
    Subject Length (words) & 5.7$\pm$0.02 & 6.9$\pm$0.01 & +1.28 \\
    Message Length (words) & 8.7$\pm$0.06 & 9.9$\pm$0.05 & +1.34 \\
    Pre-Commit Code Length (words) & 3269.9$\pm$298.8 & 59.1$\pm$0.19 & -3210.9 \\
    Post-Commit Code Length (words) & 3269.8$\pm$299.5 & 77.6$\pm$0.23 & -3214.2 \\   
    \bottomrule
    \end{tabular}
    \caption{
        \textbf{The effect of data filters on subject, message, and code lengths}. We compare differences in word statistics of \data{} and \dataft{}.
    }
    \label{tab:filterstats}
\end{table*}


\section{Comparing \data{} and The Stack}
\label{sec:stack}

In \autoref{tab:overlap} we provide statistics on repositories and usernames of \data{} and The Stack~\citep{kocetkov2022stack}. \data{} contains a total of 1,934,255 repositories. Around half (49.3\%) of them are also in The Stack. However, The Stack only provides the raw code files of these repositories from some fixed point in time. \data{} contains the changes made to the code files in the form of commits. Thus, the same code file may appear multiple times in \data{} for each change that was made to it. Therefore, The Stack only contains 3 terabytes of data, while \data{} contains close to 4.

\begin{table*}[htbp]
    \centering
    \begin{tabular}{l|cc|cc}
    \toprule
    Statistic ($\downarrow$) & \data{} & The Stack 1.2 & Shared & Shared (\%) \\
    \midrule
    Repositories & 1,934,255 & 18,712,378 & 954,135 & 49.3\% \\
    Usernames & 825,885 & 6,434,196 & 663,050 & 80.3\% \\
    \bottomrule
    \end{tabular}
    \caption{
        \textbf{Overlap in repositories and usernames of \data{} and The Stack.}
    }
    \label{tab:overlap}
\end{table*}

\FloatBarrier

\section{Pretraining on \data{}}
\label{sec:pretraining}


Due to the scale of \data{}, it is also adequate as a large-scale pretraining dataset. We have included parts of \data{} during the pretraining of StarCoder~\citep{li2023starcoder} in the format of \verb|<commit_before>code_before<commit_msg>message<commit_after>| \verb|code_after|.
We also pretrain a new model, named \modelsc{}, with the same architecture as SantaCoder~\citep{allal2023santacoder} on \data{} using this format. We filter \data{} for our six evaluation languages and samples that fit within 8192 tokens leaving us a total of 35B tokens. Following prior work~\citep{muennighoff2023scaling}, we train on this data repeated close to 4 times for a total of 131B tokens taking 14 days. Detailed hyperparameters are in \autoref{sec:hyperparameters}.

In \autoref{tab:humanevalcommit}, we benchmark StarCoder and \modelsc{} on \evalf{} using the above-detailed commit format. We find that the commit format leads to very strong performance for StarCoder often surpassing the instruction tuned \model{} from \autoref{tab:humanevalpack}. However, this pretraining format is not suitable for \evale{} limiting its universality.
For \modelsc{}, we find performance comparable to SantaCoder, including checkpoints at 131B and 236B tokens. \modelsc{} performs slightly worse on Python than SantaCoder. We hypothesize that this discrepancy is due to a \textit{multilingual tax}, as \modelsc{} needs to accommodate three additional coding languages (Go, C++ and Rust). SantaCoder has thus more capacity allocated to Python, JavaScript, and Java.

\modelsc{} may also be bottlenecked by its small model size of 1.1B parameters. More research into what exactly happens during pretraining~\citep{xia2022training,biderman2023emergent} and how to unify pretraining and instruction tuning are needed. Prior work has also found that including raw code data during pretraining benefits some natural language tasks~\citep{muennighoff2023scaling,Soldaini2024DolmaAO,Groeneveld2024OLMoAT}. Future work may consider the effects of including code commit data on natural language tasks.

\begin{table*}[htbp]
    \centering
    \resizebox{\textwidth}{!}{
    \begin{tabular}{l|cccccc|c}
    \toprule
    \textbf{Model ($\downarrow$)} & \textbf{Python} & \textbf{JavaScript} & \textbf{Java} & \textbf{Go} & \textbf{C++} & \textbf{Rust} & \textbf{Avg.} \\
    \midrule
    SantaCoder (131B tokens) Instruct Format  & 6.5 & 4.2 & 2.9 & - & - & - & - \\
    SantaCoder (236B tokens) Instruct Format  & 7.1 & 4.2 & 1.8 & - & - & - & - \\
    \modelsc{} (131B tokens) Commit Format  & 3.2 & 4.9 & 1.8 & 3.6 & 4.2 & 1.7 & 3.3 \\
    \midrule
    StarCoder Commit Format & 32.7 & 33.6 & 33.0 & 31.9 & 31.6 & 20.2 & 30.5 \\  
    \bottomrule
    \end{tabular}
    }
    \caption{
        \textbf{Zero-shot pass@1 (\%) performance on \evalf{} of pretraining experiments.}
    }
    \label{tab:humanevalcommit}
\end{table*}

\FloatBarrier

\section{\eval{} Statistics}
\label{sec:stats}

\autoref{tab:stats} displays statistics of \eval{}. Docstrings are largely the same across languages leading to similar statistics except for Rust. As Rust is already a very verbose language as seen by its maximum solution length in \autoref{tab:stats}, we do not include examples of how to call the function at the end of its docstrings (see Python docstrings with examples in e.g. \autoref{fig:missinglogic}). Rust also has type annotations for every function so providing these examples is not as needed as it is for e.g. JavaScript which lacks type annotations.

\begin{table*}[htbp]
    \centering
    \begin{tabular}{l|cccccc}
    \toprule
    \textbf{Statistic ($\downarrow$)} & \textbf{Python} & \textbf{JavaScript} & \textbf{Java} & \textbf{Go} & \textbf{C++} & \textbf{Rust} \\
    \midrule
    Docstring Avg. Length (chars) & 354 & 352 & 357 & 365 & 352 & 231 \\
    Docstring Min. Length (chars) & 56 & 56 & 56 & 56 & 56 & 23 \\
    Docstring Max. Length (chars) & 1207 & 1207 & 1211 & 1207 & 1207 & 1067 \\
    \midrule
    Solution Avg. Length (chars) & 181 & 234 & 259 & 354 & 295 & 339 \\
    Solution Min. Length (chars) & 16 & 19 & 18 & 29 & 17 & 17 \\
    Solution Max. Length (chars) & 864 & 1325 & 1399 & 1333 & 1144 & 2157 \\
    \bottomrule
    \end{tabular}
    \caption{\textbf{Statistics of \eval{} computed across the 164 samples for each language.}}
    \label{tab:stats}
\end{table*}

\FloatBarrier
\newpage

\section{Full Instruction Data Ablations}
\label{sec:ablations}

We provide tabular results of the ablations from \autoref{fig:ablations} in \autoref{tab:ablations}. We try some additional mixtures, however, none of them perform better than \dataft{} + OASST. We experiment with changing the formatting to be \verb|<commit_before>old code<commit_msg>message<commit_after>new code| for \dataft{} and \verb|<commit_before><commit_msg>input<commit_after>output| for OASST referred to as the "Formatting" ablation. We hypothesized that aligning the formatting during instruction tuning with the commit format that we used during pretraining (\autoref{sec:pretraining}) would improve performance. While it seems to improve performance for \evalf{} compared to our default formatting (see \autoref{fig:octocoder}), it reduces performance on the other tasks leading to a worse average score of 35.3 in \autoref{tab:ablations}. "Target Loss" refers to an ablation where we mask loss for inputs as is commonly done during instruction tuning~\citep{muennighoff2022crosslingual}. While this leads to the best performance on \evals{}, its average performance is worse compared to \dataft{} + OASST, where the loss is computed over the full sequence. We also perform an ablation where we manually select 1178 high-quality samples (725 from OASST and 89, 61, 86, 72, 70 and 75 from \dataft{} for Python, JavaScript, Java, Go, C++ and Rust, respectively). However, this manual selection did not outperform random selection for \model{}. It performed better for \modelx{}, however, hence we used it for \modelx{}. We hypothesize that our models could achieve significantly better performance by further improving the quality of the instruction data beyond. This may necessitate very careful human selection of samples and manual editing of the data to ensure a uniform style in the outputs. We leave such explorations to future work.

\begin{table*}[htbp]
    \centering
    \begin{tabular}{l|ccc|c}
    \toprule
    & \multicolumn{3}{c|}{\eval{} Python} & \\
    \textbf{Instruction Tuning Dataset ($\downarrow$)} & \textbf{Fix} & \textbf{Explain} & \textbf{Synthesize} & \textbf{Average} \\
    \midrule
    Without instruction tuning & 8.7 & 0.0 & 33.6 & 14.1 \\    
    \midrule
    Self-Instruct (SI) & 23.6 & 0.6 & 43.0 & 22.2 \\
    OASST & 23.1 & 34.5 & 46.4 & 34.7 \\
    SI + OASST & 24.9 & 28.7 & 46.2 & 33.3 \\
    xP3x + OASST & 28.4 & 28.4 & 45.0 & 33.9 \\
    \dataft{} + OASST & 30.4 & \textbf{35.1} & 46.2 & \textbf{37.2} \\
    \dataft{} + OASST (Formatting) & 31.1 & 28.9 & 45.8 & 35.3  \\
    \dataft{} + OASST (Target loss) & 29.8 & 31.2 & \textbf{47.8} & 36.3 \\
    \dataft{} + OASST (Manual) & 27.2 & 29.6 & 45.8 & 34.2 \\
    \dataft{} + xP3x + OASST & 30.9 & 29.5 & 45.9 & 35.4 \\
    \dataft{} + SI + xP3x + OASST & \textbf{31.4} & 33.8 & 46.0 & 37.1 \\
    \bottomrule
    \end{tabular}
    \caption{\textbf{Zero-shot pass@1 (\%) performance across the Python split of \eval{} for StarCoder instruction tuning data ablations.}}
    \label{tab:ablations}
\end{table*}

\FloatBarrier

\section{Line Diff Format for Fixing Code}
\label{sec:diffformat}

We finetune SantaCoder to experiment with different formatting strategies for fixing bugs comparing \textit{full code generation} and \textit{code diff generation}. When fixing a code bug, usually only a small part of the code needs to change. Only generating the code diff corresponding to the necessary change can make inference significantly more efficient by avoiding repeated characters in the output generation. We finetune SantaCoder on the Python, Java and JavaScript subset of \dataft{}. We exclude other languages as SantaCoder has only been pretrained on these three languages~\citep{allal2023santacoder}.

\paragraph{Commit Format} For \textit{full code generation}, we reuse the format that we employed for commits in StarCoder pretraining from \autoref{sec:pretraining}: \verb|<commit_before>code_before<commit_msg>| \verb|message<commit_after>code_after|. However, SantaCoder has not seen this format during pretraining and does not have special tokens like StarCoder for the delimiters. Thus, for SantaCoder e.g. \verb|<commit_before>| is tokenized as \verb|['<', 'commit', '_', 'before', '>']|.

\paragraph{Unified diff format} For \textit{code diff generation}, a simple solution is using the unified diff format,\footnote{\url{https://en.wikipedia.org/wiki/Diff\#Unified_format}} which is a standard way to display changes between code files in a compact and readable format \citep{lehman2022evolution,jung2021commitbert,xu2022combining,monperrus2021megadiff}. We depict an example of this format in \autoref{fig:unidiff}. However, the unified diff format still requires the model to output several unchanged lines below and after the actual modification. Thus, its efficiency gains are limited and there is still unnecessary duplication of the input.

\begin{figure}[htbp]
\centering
\begin{minipage}{0.48\textwidth}
\begin{lstlisting}[language=Python,breaklines=true,basicstyle=\ttfamily\scriptsize]
from typing import List

def has_close_elements(numbers: List[float], threshold: float) -> bool:
    for idx, elem in enumerate(numbers):
        for idx2, elem2 in enumerate(numbers):
            if idx != idx2:
                distance = elem - elem2
                if distance < threshold:
                    return True

    return False
\end{lstlisting}
\end{minipage}\hfill
\begin{minipage}{0.48\textwidth}
\begin{lstlisting}[language=Python,breaklines=true,basicstyle=\ttfamily\scriptsize]
from typing import List

def has_close_elements(numbers: List[float], threshold: float) -> bool:
    for idx, elem in enumerate(numbers):
        for idx2, elem2 in enumerate(numbers):
            if idx != idx2:
                distance = abs(elem - elem2)
                if distance < threshold:
                    return True

    return False
\end{lstlisting}
\end{minipage}
\begin{minipage}{0.6\textwidth}
\begin{lstlisting}[language=Python,breaklines=true,basicstyle=\ttfamily\scriptsize]
@@ -4,7 +4,7 @@
     for idx, elem in enumerate(numbers):
         for idx2, elem2 in enumerate(numbers):
             if idx != idx2:
-                distance = elem - elem2
+                distance = abs(elem - elem2)
                 if distance < threshold:
                     return True
\end{lstlisting}
\end{minipage}
\caption{\textbf{The first problem from the \evalf{} Python split and the necessary change to fix the bug in unified diff format}. \emph{Top:} Code with and without the bug from \autoref{fig:missinglogic}. \emph{Bottom:} Necessary change to fix the bug in unified diff format.}
\label{fig:unidiff}
\end{figure}

\begin{figure}[htbp]
\centering
\begin{lstlisting}[language=Python,breaklines=true,basicstyle=\ttfamily\scriptsize]
-  7             distance = elem - elem2
+  7             distance = abs(elem - elem2)
\end{lstlisting}
\caption{\textbf{The line diff format for the problem from \autoref{fig:unidiff}.}}
\label{fig:linediff}
\end{figure}

\paragraph{Line diff format} To address the inefficiencies of the unified diff format, we propose the line diff format for representing code differences. There are two requirements for our format: \textbf{(1)} The diff can be unambiguously applied to the code before the commit to generate the code after the commit, and \textbf{(2)} the code diff should be as short as possible to maximize efficiency by avoiding the inclusion of unchanged code. In \autoref{fig:linediff}, we show how our format addresses these. The line diff format keeps track of each change sequentially line-by-line to ensure the code can be correctly modified. By focusing only on the lines that change, we reduce the number of characters in the diff by 70\% compared to the unified diff representation in \autoref{fig:unidiff}.

Both the unified diff format and our line diff format require the model to predict line numbers. This is very challenging when training on raw code as models need to count and keep track of line numbers. To simplify line number prediction, we automatically add line numbers to the raw code in the finetuning dataset for the line diff format. This allows the model to simply copy the line number into the output simplifying the diff generation. However, it diminishes efficiency slightly by adding additional input tokens that the model needs to process.

As summarized in \autoref{tab:santacodercommit}, finetuning SantaCoder using the line diff format significantly improves performance compared to prior finetuning on \evalf{} across all languages. It also outperforms finetuning using the commit format, which only provides gains on JavaScript and Java compared to no finetuning. However, finetuning on the diff format may converge slower than the commit format as the diff format significantly differs from the raw code seen during pretraining. Figures~\ref{fig:jsexample_linediff}, \ref{fig:javaexample_linediff}, \ref{fig:pyexample_linediff} show line diff generations of our model. A limitation of our current line diff implementation is that it does not handle code insertion well. The inserted lines may change the line numbers of all following lines, which can result in problems when applying the diff. Further, the diff format is not useful for \evale{} and \evals{}. Future work could consider training models that can both be instructed to use the line diff format, such as for \evalf{}, but also explain or synthesize code without producing a diff.

\begin{table*}[htbp]
    \centering
    \begin{tabular}{l|ccc}
    \toprule
    \textbf{Model} & \textbf{Python} & \textbf{JavaScript} & \textbf{Java} \\
    \midrule
    SantaCoder & 7.1 & 4.2 & 1.8 \\
    SantaCoder + Commit format finetuning & 3.8 & 5.3 & 9.2 \\
    SantaCoder + Line diff format finetuning & \textbf{9.9} & \textbf{9.7} & \textbf{10.0} \\
    \bottomrule
    \end{tabular}
    \caption{
        \textbf{Zero-shot pass@1 (\%) performance on \evalf{} of SantaCoder formatting experiments.}
    }
    \label{tab:santacodercommit}
\end{table*}

\begin{figure}[htbp]
\centering
\begin{lstlisting}[language=Java,breaklines=true,basicstyle=\ttfamily\scriptsize]
-  3     let depth = 0, max_depth = 0;
+  3     let depth = 0, max_depth = 1;
- 12     return max_depth;
+ 12     return max_depth - 1;
- 14   return paren_string.split(' ')
- 15           .filter(x => x != '')
- 16           .map(x => parseParenGroup(x));
- 17 }
+ 14   let paren_list = paren_string.split(' ');
+ 15   let nested_parens = paren_list.map(x => parseParenGroup(x));
+ 16   return nested_parens.reduce((prev, curr) => {
+ 17     if (prev == 0) {
+ 18       return curr;
+ 19     } else {
+ 20       return curr - 1;
+ 21     }
+ 22   });
+ 23 }
\end{lstlisting}
\caption{\textbf{A line diff generation of our model on a JavaScript \evalf{} problem.}}
\label{fig:jsexample_linediff}
\end{figure}

\begin{figure}[htbp]
\centering
\begin{lstlisting}[language=Java,breaklines=true,basicstyle=\ttfamily\scriptsize]
- 18                 if (current_depth < 0) {
+ 18                 if (current_depth < 0 && current_string.length() > 0) {
\end{lstlisting}
\caption{\textbf{A line diff generation of our model on a Java \evalf{} problem.}}
\label{fig:javaexample_linediff}
\end{figure}

\begin{figure}[htbp]
\centering
\begin{lstlisting}[language=Python,breaklines=true,basicstyle=\ttfamily\scriptsize]
-  2     for i, l1 in enumerate(l):
-  3         for j in range(i, len(l)):
+  2     for i in range(0, len(l)):
+  3         for j in range(i+1, len(l)):\end{lstlisting}
\caption{\textbf{A line diff generation of our model on a Python \evalf{} problem.}}
\label{fig:pyexample_linediff}
\end{figure}

\FloatBarrier

\section{Results on \evalfd{}}
\label{sec:docstrings}

The default version of \evalf{} does not include docstrings, but only provides the unit tests to the model alongside the buggy function. An alternative is providing docstrings as the source of ground truth for the model to fix the buggy function. Solving from docstrings is generally easier for models than from tests, as models can also solve it via pure code synthesis without looking at the buggy function at all. We provide results of some models on this variant in~\autoref{tab:docs}. For StarCoder, we distinguish two prompting formats: An instruction to fix bugs like in \autoref{fig:eval} or the commit format it has seen during pretraining (\autoref{sec:pretraining}). \model{} performs strongly on this variant. However, directly using StarCoder with the commit format from pretraining (\autoref{sec:pretraining}) is slightly better. This is in line with the commit format from pretraining also performing slightly better on \evalf{} in \autoref{tab:humanevalcommit} compared to \model{} in \autoref{tab:humanevalpack}. Diff Codegen 2B~\citep{bradley2023diffmodels} performs poorly as its predicted code diffs are often irrelevant to the actual bug, see \autoref{fig:diffcodegenjava}.

\begin{table*}[htbp]
    \centering
    \begin{tabular}{l|cccccc|c}
    \toprule
    Model & Python & JavaScript & Java & Go & C++ & Rust & Avg.\\
    \midrule
    \multicolumn{7}{c}{Non-permissive models} \\
    \midrule
    GPT-4 & 88.4 & 80.5 & 82.9 & 81.1 & 82.3 & 68.9 & \underline{80.7}\\ 
    \midrule
    \multicolumn{7}{c}{Permissive Models} \\
    \midrule
    Diff Codegen 2B & 0.0 & 0.1 & 0.0 & 0.3 & 0.0 & 0.2 & 0.1 \\
    StarCoder Commit Format & \textbf{58.8} & \textbf{49.2} & 43.9 & \textbf{55.2} & \textbf{51.5} & \textbf{41.8} & \textbf{50.1} \\
    StarCoder Instruct Format & 41.7 & 30.7 & 44.3 & 34.5 & 28.7 & 14.0 & 26.5 \\  
    \model{} & 53.8 & 48.1 & \textbf{54.3} & 54.9 & 49.2 & 32.1 & 48.7 \\
    \bottomrule
    \end{tabular}
    \caption{
        \textbf{Zero-shot pass@1 (\%) performance on \evalfd{}.}
    }
    \label{tab:docs}
\end{table*}

\FloatBarrier

\section{\evalf{} Bug Types}
\label{sec:bugs}

\autoref{tab:bugtypes} contains an overview of bugs that were manually added by one of the authors to HumanEval solutions for the construction of \evalf{}. Figures~\ref{fig:missinglogic}-\ref{fig:functionmisuse} contain an example of each type from the Python split. The bug type for each problem is the same across all programming languages in \evalf{}, but for a few samples it affects a different part of the solution due to the code solutions not being perfectly parallel across languages.

\begin{table}[htbp]
\centering
\resizebox{\textwidth}{!}{
\begin{tabular}{ccccc}
\toprule
Bug type & Subtype & Explanation & Example & Count \\
\midrule
Missing logic & & Misses code needed to solve the problem & \autoref{fig:missinglogic} & 33 \\
Excess logic & & Contains excess code leading to mistakes & \autoref{fig:excesslogic} & 31 \\
\multirow{4}{*}{Wrong logic} & Value misuse & An incorrect value is used & 
\autoref{fig:valuemisuse} & 44\\
& Operator misuse & An incorrect operator is used & 
\autoref{fig:operatormisuse} & 25 \\
& Variable misuse & An incorrect variable is used & 
\autoref{fig:variablemisuse} & 23\\
& Function misuse & An incorrect function is used & 
\autoref{fig:functionmisuse} & 8\\
\midrule
Total & & & & 164 \\
\bottomrule
\end{tabular}
}
\caption{\textbf{\evalf{} bug types.}}
\label{tab:bugtypes}
\end{table}

\begin{figure}[htbp]
\centering
\begin{minipage}{0.45\textwidth}
\begin{lstlisting}[language=Python,breaklines=true,basicstyle=\ttfamily\scriptsize]
from typing import List


def has_close_elements(numbers: List[float], threshold: float) -> bool:
    """ Check if in given list of numbers, are any two numbers closer to each other than
    given threshold.
    >>> has_close_elements([1.0, 2.0, 3.0], 0.5)
    False
    >>> has_close_elements([1.0, 2.8, 3.0, 4.0, 5.0, 2.0], 0.3)
    True
    """
    for idx, elem in enumerate(numbers):
        for idx2, elem2 in enumerate(numbers):
            if idx != idx2:
                distance = abs(elem - elem2)
                if distance < threshold:
                    return True

    return False
\end{lstlisting}
\end{minipage}\hfill
\begin{minipage}{0.45\textwidth}
\begin{lstlisting}[language=Python,breaklines=true,basicstyle=\ttfamily\scriptsize]
from typing import List


def has_close_elements(numbers: List[float], threshold: float) -> bool:
    """ Check if in given list of numbers, are any two numbers closer to each other than
    given threshold.
    >>> has_close_elements([1.0, 2.0, 3.0], 0.5)
    False
    >>> has_close_elements([1.0, 2.8, 3.0, 4.0, 5.0, 2.0], 0.3)
    True
    """
    for idx, elem in enumerate(numbers):
        for idx2, elem2 in enumerate(numbers):
            if idx != idx2:
                distance = elem - elem2
                if distance < threshold:
                    return True

    return False
\end{lstlisting}
\end{minipage}
\caption{\textbf{Missing logic bug example.} The buggy code (\emph{right}) misses the 'abs' statement.}
\label{fig:missinglogic}
\end{figure}

\begin{figure}[htbp]
\centering
\begin{minipage}{0.45\textwidth}
\begin{lstlisting}[language=Python,breaklines=true,basicstyle=\ttfamily\scriptsize]
def truncate_number(number: float) -> float:
    """ Given a positive floating point number, it can be decomposed into
    and integer part (largest integer smaller than given number) and decimals
    (leftover part always smaller than 1).

    Return the decimal part of the number.
    >>> truncate_number(3.5)
    0.5
    """
    return number % 1.0
\end{lstlisting}
\end{minipage}\hfill
\begin{minipage}{0.45\textwidth}
\begin{lstlisting}[language=Python,breaklines=true,basicstyle=\ttfamily\scriptsize]
def truncate_number(number: float) -> float:
    """ Given a positive floating point number, it can be decomposed into
    and integer part (largest integer smaller than given number) and decimals
    (leftover part always smaller than 1).

    Return the decimal part of the number.
    >>> truncate_number(3.5)
    0.5
    """
    return number % 1.0 + 1.0
\end{lstlisting}
\end{minipage}
\caption{\textbf{Excess logic bug example.} The buggy code (\emph{right}) incorrectly adds 1 to the result.}
\label{fig:excesslogic}
\end{figure}

\begin{figure}[htbp]
\centering
\begin{minipage}{0.45\textwidth}
\begin{lstlisting}[language=Python,breaklines=true,basicstyle=\ttfamily\scriptsize]
from typing import List, Tuple


def sum_product(numbers: List[int]) -> Tuple[int, int]:
    """ For a given list of integers, return a tuple consisting of a sum and a product of all the integers in a list.
    Empty sum should be equal to 0 and empty product should be equal to 1.
    >>> sum_product([])
    (0, 1)
    >>> sum_product([1, 2, 3, 4])
    (10, 24)
    """
    sum_value = 0
    prod_value = 1

    for n in numbers:
        sum_value += n
        prod_value *= n
    return sum_value, prod_value
\end{lstlisting}
\end{minipage}\hfill
\begin{minipage}{0.45\textwidth}
\begin{lstlisting}[language=Python,breaklines=true,basicstyle=\ttfamily\scriptsize]
from typing import List, Tuple


def sum_product(numbers: List[int]) -> Tuple[int, int]:
    """ For a given list of integers, return a tuple consisting of a sum and a product of all the integers in a list.
    Empty sum should be equal to 0 and empty product should be equal to 1.
    >>> sum_product([])
    (0, 1)
    >>> sum_product([1, 2, 3, 4])
    (10, 24)
    """
    sum_value = 0
    prod_value = 0

    for n in numbers:
        sum_value += n
        prod_value *= n
    return sum_value, prod_value
\end{lstlisting}
\end{minipage}
\caption{\textbf{Value misuse bug example.} The buggy code (\emph{right}) incorrectly initializes the product to 0.}
\label{fig:valuemisuse}
\end{figure}

\begin{figure}[htbp]
\centering
\begin{minipage}{0.45\textwidth}
\begin{lstlisting}[language=Python,breaklines=true,basicstyle=\ttfamily\scriptsize]
from typing import List


def below_zero(operations: List[int]) -> bool:
    """ You're given a list of deposit and withdrawal operations on a bank account that starts with
    zero balance. Your task is to detect if at any point the balance of account fallls below zero, and
    at that point function should return True. Otherwise it should return False.
    >>> below_zero([1, 2, 3])
    False
    >>> below_zero([1, 2, -4, 5])
    True
    """
    balance = 0

    for op in operations:
        balance += op
        if balance < 0:
            return True

    return False
\end{lstlisting}
\end{minipage}\hfill
\begin{minipage}{0.45\textwidth}
\begin{lstlisting}[language=Python,breaklines=true,basicstyle=\ttfamily\scriptsize]
from typing import List


def below_zero(operations: List[int]) -> bool:
    """ You're given a list of deposit and withdrawal operations on a bank account that starts with
    zero balance. Your task is to detect if at any point the balance of account fallls below zero, and
    at that point function should return True. Otherwise it should return False.
    >>> below_zero([1, 2, 3])
    False
    >>> below_zero([1, 2, -4, 5])
    True
    """
    balance = 0

    for op in operations:
        balance += op
        if balance == 0:
            return True

    return False
\end{lstlisting}
\end{minipage}
\caption{\textbf{Operator misuse bug example.} The buggy code (\emph{right}) incorrectly checks for equality with 0.}
\label{fig:operatormisuse}
\end{figure}

\begin{figure}[htbp]
\centering
\begin{minipage}{0.45\textwidth}
\begin{lstlisting}[language=Python,breaklines=true,basicstyle=\ttfamily\scriptsize]
from typing import List


def mean_absolute_deviation(numbers: List[float]) -> float:
    """ For a given list of input numbers, calculate Mean Absolute Deviation
    around the mean of this dataset.
    Mean Absolute Deviation is the average absolute difference between each
    element and a centerpoint (mean in this case):
    MAD = average | x - x_mean |
    >>> mean_absolute_deviation([1.0, 2.0, 3.0, 4.0])
    1.0
    """
    mean = sum(numbers) / len(numbers)
    return sum(abs(x - mean) for x in numbers) / len(numbers)
\end{lstlisting}
\end{minipage}\hfill
\begin{minipage}{0.45\textwidth}
\begin{lstlisting}[language=Python,breaklines=true,basicstyle=\ttfamily\scriptsize]
from typing import List


def mean_absolute_deviation(numbers: List[float]) -> float:
    """ For a given list of input numbers, calculate Mean Absolute Deviation
    around the mean of this dataset.
    Mean Absolute Deviation is the average absolute difference between each
    element and a centerpoint (mean in this case):
    MAD = average | x - x_mean |
    >>> mean_absolute_deviation([1.0, 2.0, 3.0, 4.0])
    1.0
    """
    mean = sum(numbers) / len(numbers)
    return sum(abs(x - mean) for x in numbers) / mean
\end{lstlisting}
\end{minipage}
\caption{\textbf{Variable misuse bug example.} The buggy code (\emph{right}) incorrectly divides by the mean.}
\label{fig:variablemisuse}
\end{figure}

\begin{figure}[htbp]
\centering
\begin{minipage}{0.45\textwidth}
\begin{lstlisting}[language=Python,breaklines=true,basicstyle=\ttfamily\scriptsize]
def flip_case(string: str) -> str:
    """ For a given string, flip lowercase characters to uppercase and uppercase to lowercase.
    >>> flip_case('Hello')
    'hELLO'
    """
    return string.swapcase()    
\end{lstlisting}
\end{minipage}\hfill
\begin{minipage}{0.45\textwidth}
\begin{lstlisting}[language=Python,breaklines=true,basicstyle=\ttfamily\scriptsize]
def flip_case(string: str) -> str:
    """ For a given string, flip lowercase characters to uppercase and uppercase to lowercase.
    >>> flip_case('Hello')
    'hELLO'
    """
    return string.lower()    
\end{lstlisting}
\end{minipage}
\caption{\textbf{Function misuse bug example.} The buggy code (\emph{right}) incorrectly uses the 'lower()' function.}
\label{fig:functionmisuse}
\end{figure}

\FloatBarrier

\section{Performance Breakdown by \evalf{} Bug Type}
\label{sec:modelbugs}

All bugs in \evalf{} are categorized into bug types as described in \autoref{sec:bugs}. In \autoref{tab:bugtypeperf}, we break down the \evalf{} performance of select models from \autoref{tab:humanevalpack} by bug type. We find that models struggle most with bugs that require removing excess logic (e.g. \autoref{fig:excesslogic}). WizardCoder is only able to solve 11\% of excess logic bugs while solving about four times more bugs that relate to value misuse. The performance of \modelx{} and \model{} is more stable than WizardCoder across the different bug types, possibly due to the diversity of \dataft{} as displayed in \autoref{fig:data}. GPT-4 performs best across all bug types.

\begin{table}[htbp]
\centering
\resizebox{\textwidth}{!}{
\begin{tabular}{cc|cccc}
\toprule
Bug type & Subtype & \modelx{} & \model{} & WizardCoder & GPT-4 \\
\midrule
Missing logic & & 24.2 & 24.4 & 31.2 & 45.5 \\
Excess logic & & 16.3 & 16.9 & 11.0 & 38.7 \\
\multirow{4}{*}{Wrong logic} & Value misuse & 33.2 & 34.7 & 45.1 & 50.0 \\
& Operator misuse & 32.8 & 42.0 & 34.4 & 56.0 \\
& Variable misuse & 35.7 & 33.7 & 30.4 & 43.5 \\
& Function misuse & 25.0 & 37.5 & 37.5 & 50.0 \\
\midrule
\multicolumn{2}{c|}{Overall} & 28.1 & 30.4 & 31.8 & 47.0 \\
\bottomrule
\end{tabular}
}
\caption{\textbf{Breakdown of \evalf{} Python pass@1 (\%) performance by bug type for select models.} Statistics for each bug type are in \autoref{tab:bugtypes}.}
\label{tab:bugtypeperf}
\end{table}

\FloatBarrier

\section{\evale{} with Fill-In-The-Middle}
\label{sec:explainfim}

In \autoref{tab:humanevalpack}, all models are prompted with the same instruction to provide an explanation (but using slightly different formats, see \autoref{sec:prompts}). For StarCoder, we can alternatively prompt it with the Fill-in-the-Middle (FIM) technique~\citep{bavarian2022fim}, which it already knows from pretraining~\citep{li2023starcoder}. To do so, we provide it with a prompt as shown in \autoref{fig:fim} to generate the docstring. While a docstring is not necessarily an explanation, it can be similar. The results in \autoref{tab:fim} show that it performs significantly better with this prompting strategy than in \autoref{tab:humanevalpack}. However, it still falls short of \model{} and other models likely due to the imperfect approximation of an explanation.

\begin{figure}[htbp]
\hrulefill

\begin{lstlisting}[language=Python,breaklines=true,basicstyle=\ttfamily\scriptsize]
<fim_prefix>from typing import List


def has_close_elements(numbers: List[float], threshold: float) -> bool:
    """ <fim_suffix>
    """
    for idx, elem in enumerate(numbers):
        for idx2, elem2 in enumerate(numbers):
            if idx != idx2:
                distance = abs(elem - elem2)
                if distance < threshold:
                    return True

    return False<fim_middle>
\end{lstlisting}

\hrulefill

\caption{\textbf{FIM prompt example for StarCoder.}}
\label{fig:fim}
\end{figure}

\begin{table*}[htbp]
    \centering
    \begin{tabular}{l|cccccc|c}
    \toprule
    \textbf{Model} & \textbf{Python} & \textbf{JavaScript} & \textbf{Java} & \textbf{Go} & \textbf{C++} & \textbf{Rust} & \textbf{Avg.} \\
    \midrule
    StarCoder FIM & 19.4 & 17.6 & 16.3 & 11.8 & 17.9 & 16.7 & 16.6 \\
    \bottomrule
    \end{tabular}
    \caption{\textbf{Performance of StarCoder on \evale{} with FIM.}}
    \label{tab:fim}
\end{table*}

\section{\evale{} BLEU and METEOR comparison}
\label{sec:explainmetrics}

By default, we use pass@$k$ to evaluate \evale{} (\autoref{sec:eval}). In \autoref{tab:metrics} we compare it with BLEU~\citep{papineni2002bleu} and METEOR~\citep{banerjee2005meteor}. While our pass@$k$ formulation does not require a ground truth explanation, BLEU and METEOR do. This can be a limiting factor. For this evaluation, we use the function docstrings as the ground-truth explanation to compute the BLEU and METEOR scores. We compute BLEU and METEOR for each of the $n=20$ generations (\ref{sec:eval}) and select the highest score. The scores are then averaged across the 164 samples for each language. Rust scores are the highest which is likely due to Rust docstrings containing no example function calls (\autoref{sec:stats}).

\begin{table*}[htbp]
    \centering
    \begin{tabular}{l|cccccc|c}
    \toprule
    \textbf{Metric ($\downarrow$)} & \textbf{Python} & \textbf{JavaScript} & \textbf{Java} & \textbf{Go} & \textbf{C++} & \textbf{Rust} & \textbf{Avg.} \\
    \midrule
    pass@$1$ & 35.1 & 24.5 & 27.3 & 21.1 & 24.1 & 14.8 & 24.5 \\
    BLEU-1 & 7.1 & 7.0 & 6.3 & 6.1 & 6.4 & 8.1 & 6.8 \\
    BLEU-2/3/4 & 0.0 & 0.0 & 0.0 & 0.0 & 0.0 & 0.0 & 0.0 \\
    METEOR & 7.8 & 7.6 & 7.2 & 7.4 & 7.0 & 9.4 & 7.7 \\
    \bottomrule
    \end{tabular}
    \caption{\textbf{Comparison of different metrics for \evale{} on explanations by \model{}.} Pass@$1$ is computed with respect to a generated solution given the explanation (\autoref{sec:eval}) while BLEU and METEOR are computed by comparing the explanation with the docstring.}
    \label{tab:metrics}
\end{table*}

\FloatBarrier

\section{Hyperparameters}
\label{sec:hyperparameters}

\paragraph{StarCoder finetuning (\model{})} For all experiments finetuning StarCoder, we use a learning rate of 5e-4 with a cosine schedule and linear warmup. We use a batch size of 32 and train for up to one epoch, as we did not observe benefits from more steps. \model{} was trained for 35 steps with a sequence length of 2048 and packing corresponding to 2.2 million total finetuning tokens. We use LoRA~\citep{hu2021lora} as we did not observe a significant difference from full finetuning. Follow-up work has further investigated this choice~\citep{zhuo2024astraios}.

\paragraph{CodeGeeX finetuning (\modelx{})} To create \modelx{}, we finetune CodeGeeX2 for 35 steps with a batch size of 48 and a learning rate of 5e-5 largely following the \model{} setup.

\paragraph{SantaCoder finetuning} For all experiments finetuning SantaCoder, we use a learning rate of 5e-5 with a cosine schedule and linear warmup. We finetune SantaCoder using a batch size of 64 for up to 200,000 steps.

\paragraph{SantaCoder pretraining (\modelsc{})} We follow the setup from \citet{allal2023santacoder} to pretrain on \data{} except for using a sequence length of 8192 and the StarCoder tokenizer, which has special tokens for the commit format delimiters (see \autoref{sec:pretraining}). \modelsc{} utilizes Multi Query Attention (MQA) \citep{shazeer2019mqa} but removes Fill-in-the-Middle (FIM) \citep{bavarian2022fim}. We conducted pretraining on 32 A100 GPUs, totaling 250k training steps, with a global batch size of 64. Other hyperparameter settings follow SantaCoder, including using Adam with $\beta_1=0.9, \beta_2=0.95, \epsilon=10^{-8}$, and a weight decay of 0.1. The learning rate is set to $2 \times 10^{-4}$ and follows a cosine decay after warming up for 2\% of the training steps. 


\section{Prompts}
\label{sec:prompts}

The prompting format can significantly impact performance. In the spirit of true few-shot learning~\citep{perez2021true} we do not optimize prompts and go with the format provided by the respective model authors or the most intuitive format if none is provided. For each task, we define an instruction, an optional context and an optional function start (\autoref{tab:prompts}). The function start is provided to make sure the model directly completes the function without having to search for the function in the model output. These three parts are then combined in slightly different ways for each model (Figures~\ref{fig:octocoder}-\ref{fig:gpt4funcstart}). We implement our evaluation using open-source frameworks~\citep{bigcode-evaluation-harness,eval-harness}.

\begin{table}[htbp]
\begin{tabular}{|p{2.5cm}|p{10.5cm}|}
\specialrule{1.5pt}{1pt}{1pt}
\multicolumn{2}{|c|}{\textbf{\evalf{}}} \\
\midrule
Instruction & Fix bugs in has\_close\_elements. \\
\midrule
Context&
\begin{lstlisting}[language=Python,breaklines=true,basicstyle=\ttfamily\scriptsize,aboveskip=-0.8 \baselineskip,belowskip=-1.5 \baselineskip]
from typing import List


def has_close_elements(numbers: List[float], threshold: float) -> bool:
    for idx, elem in enumerate(numbers):
        for idx2, elem2 in enumerate(numbers):
            if idx != idx2:
                distance = elem - elem2
                if distance < threshold:
                    return True

    return False
\end{lstlisting} \\
\midrule
Function start &
\begin{lstlisting}[language=Python,breaklines=true,basicstyle=\ttfamily\scriptsize,aboveskip=-0.8 \baselineskip,belowskip=-1.5 \baselineskip]
from typing import List


def has_close_elements(numbers: List[float], threshold: float) -> bool:
\end{lstlisting} \\
\specialrule{1.5pt}{1pt}{1pt}
\multicolumn{2}{|c|}{\textbf{\evale{}}} \\
\midrule
Instruction 

(Describe) & Provide a concise natural language description of the code using at most 213 characters. \\
\midrule
Context 

(Describe) &
\begin{lstlisting}[language=Python,breaklines=true,basicstyle=\ttfamily\scriptsize,aboveskip=-0.8 \baselineskip,belowskip=-1.5 \baselineskip]
from typing import List


def has_close_elements(numbers: List[float], threshold: float) -> bool:
    for idx, elem in enumerate(numbers):
        for idx2, elem2 in enumerate(numbers):
            if idx != idx2:
                distance = abs(elem - elem2)
                if distance < threshold:
                    return True

    return False
\end{lstlisting} \\
\midrule
Instruction 

(Synthesize) & Write functional code in Python according to the description. \\
\midrule
Context 

(Synthesize) & \{Description generated by the model\} \\
\midrule
Function start 

(Synthesize) &
\begin{lstlisting}[language=Python,breaklines=true,basicstyle=\ttfamily\scriptsize,aboveskip=-0.8 \baselineskip,belowskip=-1.5 \baselineskip]
from typing import List


def has_close_elements(numbers: List[float], threshold: float) -> bool:
\end{lstlisting} \\
\specialrule{1.5pt}{1pt}{1pt}
\multicolumn{2}{|c|}{\textbf{\evals{}}} \\
\midrule
Instruction & Write a Python function `has\_close\_elements(numbers: List[float], threshold: float) -> bool` to solve the following problem:

Check if in given list of numbers, are any two numbers closer to each other than
given threshold.

>>> has\_close\_elements([1.0, 2.0, 3.0], 0.5)

False

>>> has\_close\_elements([1.0, 2.8, 3.0, 4.0, 5.0, 2.0], 0.3)

True \\
\midrule
Function start &
\begin{lstlisting}[language=Python,breaklines=true,basicstyle=\ttfamily\scriptsize,aboveskip=-0.8 \baselineskip,belowskip=-1.5 \baselineskip]
from typing import List


def has_close_elements(numbers: List[float], threshold: float) -> bool:
    """ Check if in given list of numbers, are any two numbers closer to each other than given threshold.
    >>> has_close_elements([1.0, 2.0, 3.0], 0.5)
    False
    >>> has_close_elements([1.0, 2.8, 3.0, 4.0, 5.0, 2.0], 0.3)
    True
    """
\end{lstlisting} \\
\bottomrule
\end{tabular}
\caption{\textbf{Instructions and function examples used.} If no function start or no context is present, that part is not added to the prompt (and the preceding newline is also removed).}
\label{tab:prompts}
\end{table}

\FloatBarrier

\begin{figure}[htbp]
\hrulefill

Question: \{instruction\}\\\{context\}\\\\Answer:\\\{function\_start\}

\hrulefill
\caption[]{\textbf{\model{} and \modelx{} prompting format}}
\label{fig:octocoder}
\end{figure}

\begin{figure}[htbp]
\hrulefill

Below is an instruction that describes a task. Write a response that appropriately completes the request.\\\\\#\#\# Instruction:\\\{instruction\}\\\{context\}\\\\\#\#\# Response:\\\{function\_start\}

\hrulefill
\caption[]{\textbf{WizardCoder prompting format from their codebase.\footnotemark}}
\label{fig:wizardcoder}
\end{figure}
\footnotetext{\url{https://github.com/nlpxucan/WizardLM/blob/9c6fb34d012d60dc4f31348ee0a8e35335c04564/WizardCoder/src/humaneval_gen.py\#L38}}

\begin{figure}[htbp]
\hrulefill

Below is an instruction that describes a task. Write a response that appropriately completes the request.\\\\\#\#\# Instruction:\\\{instruction\}\\\{context\}\\\\\#\#\# Response:\{function\_start\}

\hrulefill
\caption[]{\textbf{InstructCodeT5+ decoder prompting format from their codebase.\footnotemark} The encoder receives the same prompt without the function start.}
\label{fig:instructcodet5p}
\end{figure}
\footnotetext{\url{https://github.com/salesforce/CodeT5/blob/main/CodeT5\%2B/humaneval/generate_codet5p.py\#L89}}

\begin{figure}[htbp]
\hrulefill

<|system|>\\<|end|>\\<|user|>\\\{instruction\}\\\{context\}<|end|>\\<|assistant|>\\\{function\_start\}

\hrulefill
\caption[]{\textbf{StarChat-$\beta$ prompting format from their documentation.\footnotemark}}
\label{fig:starchatbeta}
\end{figure}
\footnotetext{\url{https://huggingface.co/HuggingFaceH4/starchat-beta}}

\begin{figure}[htbp]
\hrulefill

\{context\}\\\{instruction\}\\\{function\_start\}

\hrulefill
\caption[]{\textbf{Default prompting format (StarCoder, BLOOMZ).} Used for \autoref{fig:eval}.}
\label{fig:default}
\end{figure}

\begin{figure}[htbp]
\hrulefill

\{context\}\\\{instruction\}

\hrulefill
\caption[]{\textbf{GPT-4 prompting format.} Same as \autoref{fig:default} but excluding the function start, as we found the model would not continue from the provided function start, likely due to the chat mode.}
\label{fig:gpt4}
\end{figure}

\begin{figure}[htbp]
\hrulefill

\{context\}\\\{instruction\}\\Start your code with:\\\{func\_start\}

\hrulefill
\caption[]{\textbf{GPT-4 prompting format for \evale{} (Synthesize).} We include the function start for the synthesize part of \evale{} in case the function name is not included in the code description generated by GPT-4.}
\label{fig:gpt4funcstart}
\end{figure}

\begin{figure}[htbp]
\hrulefill

<NME> \{filename\}\\<BEF> \{context\}\\<MSG> \{instruction\}\\<DFF>

\hrulefill
\caption[]{\textbf{Prompting format for Diff Codegen 2B evaluated in \autoref{sec:docstrings}.}}
\label{fig:diff}
\end{figure}

\FloatBarrier

\section{Examples}
\label{sec:examples}

\subsection{\model{}}
\label{sec:examplesmodel}

\begin{figure}[htbp]
\hrulefill

\emph{Input:}

\hrulefill

Question: Fix bugs in sum\_product.
\begin{lstlisting}[language=C++,breaklines=true,basicstyle=\ttfamily\scriptsize]
#include<stdio.h>
#include<vector>
using namespace std;
#include<algorithm>
#include<math.h>
#include<stdlib.h>
vector<int> sum_product(vector<int> numbers){
    int sum=0,product=0;
    for (int i=0;i<numbers.size();i++)
    {
        sum+=numbers[i];
        product*=numbers[i];
    }
    return {sum,product};
}
\end{lstlisting}

Answer:
\begin{lstlisting}[language=C++,breaklines=true,basicstyle=\ttfamily\scriptsize]
#include<stdio.h>
#include<vector>
using namespace std;
#include<algorithm>
#include<math.h>
#include<stdlib.h>
vector<int> sum_product(vector<int> numbers){
\end{lstlisting}

\hrulefill

\model{}:

\hrulefill

\begin{lstlisting}[language=C++,breaklines=true,basicstyle=\ttfamily\scriptsize]
    int sum=0,product=1;
    for (int i=0;i<numbers.size();i++)
    {
        sum+=numbers[i];
        product*=numbers[i];
    }
    return {sum,product};
}
\end{lstlisting}

\hrulefill

\caption{\textbf{Correct solution from \model{} to a \evalf{} C++ problem.} The corresponding Python problem is depicted in \autoref{fig:valuemisuse}.}
\label{fig:modelcpp}
\end{figure}

\begin{figure}[htbp]
\hrulefill

\emph{Input:}

\hrulefill

Question: Write a Java function `public String decodeShift(String s)` to solve the following problem:
takes as input string encoded with encodeShift function. Returns decoded string.

Answer: 
\begin{lstlisting}[language=Java,breaklines=true,basicstyle=\ttfamily\scriptsize]
import java.util.*;
import java.lang.*;

class Solution {
    /**
    returns encoded string by shifting every character by 5 in the alphabet.
     */
    public String encodeShift(String s) {
        StringBuilder sb = new StringBuilder();
        for (char ch : s.toCharArray()) {
            sb.append((char) ('a' + ((ch + 5 - 'a') % 26)));
        }
        return sb.toString();
    }

    /**
    takes as input string encoded with encodeShift function. Returns decoded string.
     */
    public String decodeShift(String s) {
\end{lstlisting}

\hrulefill

\emph{\model{}:}

\hrulefill

\begin{lstlisting}[language=Java,breaklines=true,basicstyle=\ttfamily\scriptsize]
        StringBuilder sb = new StringBuilder();
        for (char ch : s.toCharArray()) {
            sb.append((char) ('a' + ((ch - 'a' - 5) % 26)));
        }
        return sb.toString();
    }
}
\end{lstlisting}
\caption{\textbf{Incorrect solution from \model{} to a \evals{} Java problem.} The correct solution would be to add 21 to ch instead of subtracting 5 from it.}
\label{fig:modeljava}
\end{figure}

\begin{figure}[htbp]
\hrulefill

\emph{Input:}

\hrulefill

Question: Provide a concise natural language description of the code using at most 671 characters.
\begin{lstlisting}[language=Go,breaklines=true,basicstyle=\ttfamily\scriptsize]
func FruitDistribution(s string,n int) int {
    split := strings.Split(s, " ")
        for _, i := range split {
                atoi, err := strconv.Atoi(i)
                if err != nil {
                        continue
                }
                n = n - atoi
        }
        return n
}
\end{lstlisting}

Answer: 

\hrulefill

\emph{\model{}:}

\hrulefill

The function FruitDistribution takes two arguments, a string s and an integer n. The function splits the string s into a slice of strings using the space character as the delimiter. The function iterates over the slice and converts each string to an integer using the strconv.Atoi function. If the conversion is successful, the integer is subtracted from n. The function returns the value of n after all conversions have been performed.

\hrulefill

\emph{Input (without previous input):}

\hrulefill

Question: Write functional code in Go according to the description.

The function FruitDistribution takes two arguments, a string s and an integer n. The function splits the string s into a slice of strings using the space character as the delimiter. The function iterates over the slice and converts each string to an integer using the strconv.Atoi function. If the conversion is successful, the integer is subtracted from n. The function returns the value of n after all conversions have been performed.

Answer:
\begin{lstlisting}[language=Go,breaklines=true,basicstyle=\ttfamily\scriptsize]
func FruitDistribution(s string,n int) int {
\end{lstlisting}

\hrulefill

\emph{\model{}:}

\hrulefill

\begin{lstlisting}[language=Go,breaklines=true,basicstyle=\ttfamily\scriptsize]
	fruits := strings.Split(s, " ")
	for _, fruit := range fruits {
		if num, err := strconv.Atoi(fruit); err == nil {
			n -= num
		}
	}
	return n
}
\end{lstlisting}

\hrulefill

\caption{\textbf{Correct solution from \model{} to a \evale{} Go problem.} Given its own description \model{} writes a correct solution to the problem. Its solution is more compact and efficient than the original solution.}
\label{fig:modelgo}
\end{figure}



\FloatBarrier

\subsection{GPT-4}
\label{sec:examplesgpt4}

\begin{figure}[htbp]
\hrulefill

\emph{User:}

\hrulefill

Write a Go function `func RescaleToUnit(numbers []float64) []float64` to solve the following problem:

Given a list of numbers (of at least two elements), apply a linear transform to that list,

such that the smallest number will become 0 and the largest will become 1

\verb|>>>| RescaleToUnit([1.0, 2.0, 3.0, 4.0, 5.0])

[0.0, 0.25, 0.5, 0.75, 1.0]

\hrulefill

\emph{GPT-4:}

\hrulefill

Here is a Go function that solves the problem:
\begin{lstlisting}[language=Go,breaklines=true]
package main

import (
    "fmt"
    "math"
)

func RescaleToUnit(numbers []float64) []float64 {
    min := numbers[0]
    max := numbers[0]
    
    // Find the min and max values in the list
    for _, number := range numbers {
        if number < min {
            min = number
        }
        if number > max {
            max = number
        }
    }
    
    // Apply the linear transform to each number in the list
    for i, number := range numbers {
        numbers[i] = (number - min) / (max - min)
    }
    
    return numbers
}

func main() {
    fmt.Println(RescaleToUnit([]float64{1.0, 2.0, 3.0, 4.0, 5.0}))
}
\end{lstlisting}
This function first finds the minimum and maximum values in the list of numbers. Then it applies the linear transformation to each number in the list. The transformation is `(number - min) / (max - min)`, which scales the number such that `min` becomes 0 and `max` becomes 1. The transformed numbers are returned in a new list.

\hrulefill
\caption{\textbf{Incorrect solution from GPT-4 to a \evals{} Go problem.} GPT-4 includes an import that it does not use (math) which leads to a compilation error in Golang and thus the function does not pass the test cases.}
\label{fig:gpt4go}
\end{figure}

\begin{figure}[htbp]
\hrulefill

\emph{User:}

\hrulefill

Provide a concise natural language description of the code using at most 246 characters.
\begin{lstlisting}[language=Python,breaklines=true,basicstyle=\ttfamily\scriptsize]
def truncate_number(number: float) -> float:
    return number % 1.0    
\end{lstlisting}

\hrulefill

\emph{GPT-4:}

\hrulefill

The code defines a function that takes a float as input and returns its fractional part by using the modulus operator with 1.0.

\hrulefill

\emph{User (without previous conversation history):}

\hrulefill

The code defines a function that takes a float as input and returns its fractional part by using the modulus operator with 1.0.

Write functional code in Python according to the description.

Start your code with:
\begin{lstlisting}[language=Python,breaklines=true,basicstyle=\ttfamily\scriptsize]
def truncate_number(number: float) -> float:
\end{lstlisting}

\hrulefill

\emph{GPT-4:}

\hrulefill

\begin{lstlisting}[language=Python,breaklines=true,basicstyle=\ttfamily\scriptsize]
def truncate_number(number: float) -> float:
    return number % 1.0    
\end{lstlisting}

\hrulefill
\caption{\textbf{Correct solution from GPT-4 to a \evale{} Python problem.} GPT-4 writes a description that is significantly shorter than the docstring and thus within the character limit. It then generates the perfect solution from its own description. The docstring for this problem can be seen in \autoref{fig:excesslogic}.}
\label{fig:gpt4py}
\end{figure}

\begin{figure}[htbp]
\hrulefill

\emph{User:}

\hrulefill

\begin{lstlisting}[language=Rust,breaklines=true,basicstyle=\ttfamily\scriptsize]
fn has_close_elements(numbers:Vec<f32>, threshold: f32) -> bool{
    for i in 0..numbers.len(){
        for j in 1..numbers.len(){

            if i != j {
                let distance:f32 = numbers[i] - numbers[j];

            if distance < threshold{
                return true;
            }

            }
            
        }
    }

    return false;

}





def check(has_close_elements):
    assert has_close_elements([1.0, 2.0, 3.9, 4.0, 5.0, 2.2], 0.3) == True
    assert has_close_elements([1.0, 2.0, 3.9, 4.0, 5.0, 2.2], 0.05) == False
    assert has_close_elements([1.0, 2.0, 5.9, 4.0, 5.0], 0.95) == True
    assert has_close_elements([1.0, 2.0, 5.9, 4.0, 5.0], 0.8) == False
    assert has_close_elements([1.0, 2.0, 3.0, 4.0, 5.0, 2.0], 0.1) == True
    assert has_close_elements([1.1, 2.2, 3.1, 4.1, 5.1], 1.0) == True
    assert has_close_elements([1.1, 2.2, 3.1, 4.1, 5.1], 0.5) == False

check(has_close_elements)
\end{lstlisting}
Fix bugs in has\_close\_elements.

\hrulefill

\emph{GPT-4:}

\hrulefill

The function has\_close\_elements is supposed to check if there are any two elements in the vector that are closer than a given threshold. However, there are a few issues with the current implementation:

1. The distance calculation is not correct. It should be the absolute difference between the two numbers, not just the difference.

2. The second loop should start from i+1, not 1. This is because we only need to compare each number with the numbers after it, not the numbers before it.

Here is the corrected function:

\begin{lstlisting}[language=Rust,breaklines=true,basicstyle=\ttfamily\scriptsize]
fn has_close_elements(numbers:Vec<f32>, threshold: f32) -> bool{
    for i in 0..numbers.len(){
        for j in i+1..numbers.len(){
            let distance:f32 = (numbers[i] - numbers[j]).abs();
            if distance < threshold{
                return true;
            }
        }
    }
    return false;
}
\end{lstlisting}

This function will now correctly check if there are any two elements in the vector that are closer than the given threshold.

\hrulefill
\caption{\textbf{Correct solution from GPT-4 to a \evalf{} Rust problem.} GPT-4 correctly fixes the bug by adding the necessary 'abs' statement to compute the absolute value. It additionally improves the efficiency of the function by starting the inner loop from 'i+1' instead of 1 like in the original solution. As some problems in \eval{} are not perfectly efficient, evaluating the ability of models to improve the efficiency of the solutions is another interesting benchmarking direction \citep{yetistiren2022assessing}.}
\label{fig:gpt4rs}
\end{figure}

\begin{figure}[htbp]
\hrulefill

\emph{User:}

\hrulefill

\begin{lstlisting}[language=Python,breaklines=true,basicstyle=\ttfamily\scriptsize]
def car_race_collision(n: int):
    return n**3





def check(car_race_collision):
    assert car_race_collision(2) == 4
    assert car_race_collision(3) == 9
    assert car_race_collision(4) == 16
    assert car_race_collision(8) == 64
    assert car_race_collision(10) == 100

check(car_race_collision)
\end{lstlisting}
Fix bugs in car\_race\_collision.

\hrulefill

\emph{GPT-4:}

\hrulefill

The function car\_race\_collision seems to be working fine as it is. It takes an integer 'n' as input and returns the cube of 'n'. There don't seem to be any bugs in this function. \\\\
However, if the function is supposed to calculate something specific related to a car race collision and it's not doing that correctly, we would need more information about the expected behavior to fix it.

\hrulefill
\caption{\textbf{Incorrect solution from GPT-4 to a \evalf{} Python problem.} GPT-4 does not catch that the test cases expect the function to square, not cube n. \model{} correctly modifies the 3 to 2 in 6 out of $n=20$ generations.}
\label{fig:gpt4pyrefuse}
\end{figure}

\begin{figure}[htbp]
\hrulefill

\emph{User:}

\hrulefill

\begin{lstlisting}[language=Python,breaklines=true,basicstyle=\ttfamily\scriptsize]
def rounded_avg(n, m):
    if m < n:
        return -1
    summation = 0
    for i in range(n, m+1):
        summation += i
    return bin(round(summation/(m - n)))

def check(rounded_avg):

    # Check some simple cases
    assert rounded_avg(1, 5) == "0b11"
    assert rounded_avg(7, 13) == "0b1010"
    assert rounded_avg(964,977) == "0b1111001010"
    assert rounded_avg(996,997) == "0b1111100100"
    assert rounded_avg(560,851) == "0b1011000010"
    assert rounded_avg(185,546) == "0b101101110"
    assert rounded_avg(362,496) == "0b110101101"
    assert rounded_avg(350,902) == "0b1001110010"
    assert rounded_avg(197,233) == "0b11010111"


    # Check some edge cases that are easy to work out by hand.
    assert rounded_avg(7, 5) == -1
    assert rounded_avg(5, 1) == -1
    assert rounded_avg(5, 5) == "0b101"

check(rounded_avg)
\end{lstlisting}
Fix bugs in rounded\_avg.

\hrulefill

\emph{GPT-4:}

\hrulefill

The function rounded\_avg is supposed to calculate the average of all numbers between n and m (inclusive), round it, and then return the binary representation of the result.

There are several bugs in the function:

1. The function should return -1 if m < n, but it should also return -1 if m == n because in this case, the denominator of the average calculation would be 0, which would cause a division by zero error.

2. The function calculates the average by summing all numbers between n and m and then dividing by (m - n). However, this is incorrect because the number of integers between n and m (inclusive) is (m - n + 1), not (m - n).

3. The function uses the bin function to convert the result to binary. However, the bin function returns a string that starts with '0b' to indicate that it is a binary number. If the function is supposed to return the binary representation as an integer, it should remove this prefix.

Here is the corrected function:

\begin{lstlisting}[language=Python,breaklines=true,basicstyle=\ttfamily\scriptsize]
def rounded_avg(n, m):
    if m <= n:
        return -1
    summation = 0
    for i in range(n, m+1):
        summation += i
    average = round(summation / (m - n + 1))
    return int(bin(average)[2:])
\end{lstlisting}

This function now correctly calculates the average, rounds it, and returns the binary representation as an integer. If n is greater than or equal to m, it returns -1.

\hrulefill

\caption{\textbf{Incorrect solution from GPT-4 to a \evalf{} Python problem.} The bug consists of a "+1" missing after "m-n". GPT-4 correctly fixes that, however, it introduces two new bugs: Retuning "-1" when m=n and removing the "0b", both of which evidently lead to the provided unit tests failing.}
\label{fig:gpt4pyaddbugs}
\end{figure}

\FloatBarrier

\subsection{WizardCoder}
\label{sec:exampleswizardcoder}

\begin{figure}[htbp]
\hrulefill

\emph{Input:}

\hrulefill

Below is an instruction that describes a task. Write a response that appropriately completes the request.

\#\#\# Instruction: Fix bugs in bf.
\begin{lstlisting}[language=Python,breaklines=true,basicstyle=\ttfamily\scriptsize]
def bf(planet1, planet2):
    planet_names = ("Mercury", "Venus", "Earth", "Mars", "Jupyter", "Saturn", "Uranus", "Neptune")
    if planet1 not in planet_names or planet2 not in planet_names or planet1 == planet2:
        return ()
    planet1_index = planet_names.index(planet1)
    planet2_index = planet_names.index(planet2)
    if planet1_index < planet2_index:
        return (planet_names[planet1_index + 1: planet2_index])
    else:
        return (planet_names[planet2_index + 1 : planet1_index])

def check(bf):

    # Check some simple cases
    assert bf("Jupiter", "Neptune") == ("Saturn", "Uranus"), "First test error: " + str(len(bf("Jupiter", "Neptune")))      
    assert bf("Earth", "Mercury") == ("Venus",), "Second test error: " + str(bf("Earth", "Mercury"))  
    assert bf("Mercury", "Uranus") == ("Venus", "Earth", "Mars", "Jupiter", "Saturn"), "Third test error: " + str(bf("Mercury", "Uranus"))      
    assert bf("Neptune", "Venus") == ("Earth", "Mars", "Jupiter", "Saturn", "Uranus"), "Fourth test error: " + str(bf("Neptune", "Venus"))  


    # Check some edge cases that are easy to work out by hand.
    assert bf("Earth", "Earth") == ()
    assert bf("Mars", "Earth") == ()
    assert bf("Jupiter", "Makemake") == ()

check(bf)
\end{lstlisting}
\#\#\# Response:
\begin{lstlisting}[language=Python,breaklines=true,basicstyle=\ttfamily\scriptsize]
def bf(planet1, planet2):
\end{lstlisting}

\hrulefill

\emph{WizardCoder:}

\hrulefill

\begin{lstlisting}[language=Python,breaklines=true,basicstyle=\ttfamily\scriptsize]
    planet_names = ("Mercury", "Venus", "Earth", "Mars", "Jupyter", "Saturn", "Uranus", "Neptune")
    if planet1 not in planet_names or planet2 not in planet_names or planet1 == planet2:
        return ()
    planet1_index = planet_names.index(planet1)
    planet2_index = planet_names.index(planet2)
    if planet1_index < planet2_index:
        return (planet_names[planet1_index + 1: planet2_index])
    else:
        return (planet_names[planet2_index + 1 : planet1_index])
\end{lstlisting}

\hrulefill
\caption{\textbf{Incorrect solution from WizardCoder to a \evalf{} Python problem.} The bug consists of "Jupiter" being misspelled as "Jupyter" in the function. WizardCoder returns the same function without making a single change leaving the bug. This is a particularly difficult bug that only GPT-4 and \model{} (only for 1 out of the n=20 generations) are able to solve. }
\label{fig:wizardcoderpy}
\end{figure}

\FloatBarrier
\newpage

\subsection{BLOOMZ}
\label{sec:examplesbloomz}

\begin{figure}[htbp]
\hrulefill

\emph{Input:}

\hrulefill

\begin{lstlisting}[language=JavaScript,breaklines=true,basicstyle=\ttfamily\scriptsize]
const generateIntegers = (a, b) => {
  if (a > b) {
    let tmp = a;
    a = b;
    b = tmp;
  }
  let y = []
  for (let i = a; i <= b; i++) {
    if (i == 2 || i == 4 || i == 6 || i == 8) { y.push(i) }
  }
  return y
}
\end{lstlisting}

Provide a concise natural language description of the code using at most 218 characters.

\hrulefill

\emph{BLOOMZ:}

\hrulefill

Generate a list of integers from a to b.

\hrulefill

\emph{Input (without previous input):}

\hrulefill

Generate a list of integers from a to b.

Write functional code in JavaScript according to the description.
\begin{lstlisting}[language=JavaScript,breaklines=true,basicstyle=\ttfamily\scriptsize]
const generateIntegers = (a, b) => {
\end{lstlisting}

\hrulefill

\emph{BLOOMZ:}

\hrulefill

\begin{lstlisting}[language=JavaScript,breaklines=true,basicstyle=\ttfamily\scriptsize]
  let integers = [];
  for (let i = a; i <= b; i++) {
    integers.push(i);
  }
  return integers;
}
\end{lstlisting}

\hrulefill
\caption{\textbf{Incorrect solution from BLOOMZ to a \evale{} JavaScript problem.} BLOOMZ misses in its description that the function generates only the even digits between a and b, in ascending order. Thus, its synthesized function based on its own description simply collects all integers between a and b. BLOOMZ likely left out this part due to its bias for short and concise answers. However, this bias backfires for this task.}
\label{fig:bloomzjs}
\end{figure}

\FloatBarrier

\subsection{StarCoder}
\label{sec:examplestarcoder}

\begin{figure}[htbp]
\hrulefill

\emph{Input:}

\hrulefill

\begin{lstlisting}[language=Java,breaklines=true,basicstyle=\ttfamily\scriptsize]
import java.util.*;
import java.lang.*;

class Solution {
    public Map<String, Integer> histogram(String test) {
        Map<String, Integer> dict1 = new HashMap<>();
        List<String> list1 = Arrays.asList(test.split(" " ));
        int t = 0;
        for (String i : list1) {
            if (Collections.frequency(list1, i) > t && !i.isEmpty()) {
                t = Collections.frequency(list1, i);
            }
        }
        if (t > 0) {
            for (String i : list1) {
                if (Collections.frequency(list1, i) == t) {
                    dict1.put(i, t);
                }
            }
        }
        return dict1;
    }
}
\end{lstlisting}

Provide a concise natural language description of the code using at most 423 characters.

\hrulefill

\emph{StarCoder:}

\hrulefill

\begin{lstlisting}[language=Java,breaklines=true,basicstyle=\ttfamily\scriptsize]
public class Solution {
    public Map<String, Integer> histogram(String test) {
        
            
            int t = 0;
    
            }   
        }
        if (t > 0) {
\end{lstlisting}
Write functional code in Java according to the description.

\hrulefill

\begin{lstlisting}[language=Java,breaklines=true,basicstyle=\ttfamily\scriptsize]

import java.util.*;
import java.lang.*;

class Solution {
    public Map<String, Integer> histogram(String test) {
\end{lstlisting}

\hrulefill

\emph{StarCoder:}

\hrulefill

\begin{lstlisting}[language=Java,breaklines=true,basicstyle=\ttfamily\scriptsize]
        Map<String, Integer> map = new HashMap<String, Integer>();
        int t = 0;
        for (int i = 0; i < test.length(); i++) {
            if (test.charAt(i) == 't') {
                t++;
            }
        }
        map.put("t", t);
        return map;
    }
}
\end{lstlisting}

\hrulefill

\caption{\textbf{Incorrect solution from StarCoder to a \evale{} Java problem.} StarCoder is not capable of generating natural language descriptions and instead copies the solution. However, our post-processing removes substrings from the solution (\autoref{sec:eval}) leading to a useless description. Given this description it generates an irrelevant function. As StarCoder cannot generate natural language explanations, we disqualify it from \evale{} setting its score to 0 in \autoref{tab:humanevalpack}.}
\label{fig:starcoderjava}
\end{figure}

\FloatBarrier

\subsection{InstructCodeT5+}
\label{sec:examplesinstructcodet5p}

\begin{figure}[htbp]
\hrulefill

\emph{Input:}

\hrulefill

Below is an instruction that describes a task. Write a response that appropriately completes the request.

\#\#\# Instruction: Fix bugs in has\_close\_elements.
\begin{lstlisting}[language=Python,breaklines=true,basicstyle=\ttfamily\scriptsize]
from typing import List


def has_close_elements(numbers: List[float], threshold: float) -> bool:
    for idx, elem in enumerate(numbers):
        for idx2, elem2 in enumerate(numbers):
            if idx != idx2:
                distance = elem - elem2
                if distance < threshold:
                    return True

    return False




def check(has_close_elements):
    assert has_close_elements([1.0, 2.0, 3.9, 4.0, 5.0, 2.2], 0.3) == True
    assert has_close_elements([1.0, 2.0, 3.9, 4.0, 5.0, 2.2], 0.05) == False
    assert has_close_elements([1.0, 2.0, 5.9, 4.0, 5.0], 0.95) == True
    assert has_close_elements([1.0, 2.0, 5.9, 4.0, 5.0], 0.8) == False
    assert has_close_elements([1.0, 2.0, 3.0, 4.0, 5.0, 2.0], 0.1) == True
    assert has_close_elements([1.1, 2.2, 3.1, 4.1, 5.1], 1.0) == True
    assert has_close_elements([1.1, 2.2, 3.1, 4.1, 5.1], 0.5) == False

check(has_close_elements)
\end{lstlisting}
\#\#\# Response:from typing import List \begin{lstlisting}[language=Python,breaklines=true,basicstyle=\ttfamily\scriptsize]

def has_close_elements(numbers: List[float], threshold: float) -> bool:
\end{lstlisting}

\hrulefill

\emph{InstructCodeT5+:}

\hrulefill

\begin{lstlisting}[language=Python,breaklines=true,basicstyle=\ttfamily\scriptsize]
    for idx, elem in enumerate(numbers):
        for idx2, elem2 in enumerate(numbers):
            if idx != idx2:
                distance = elem - elem2
                if distance < threshold:
                    return True

    return False
\end{lstlisting}
\caption{\textbf{Incorrect solution from InstructCodeT5+ to a \evalf{} Python problem.} InstructCodeT5+ simply repeated the function without adding the missing "abs" statement to turn the distance into an absolute value. This is one of the simplest bugs that most other models are able to fix. For InstructCodeT5+ the prompt starts directly after "Response:" without a newline (\autoref{fig:instructcodet5p}).}
\label{fig:instructcodet5ppy}
\end{figure}

\subsection{StarChat-\texorpdfstring{$\beta$}{beta}}
\label{sec:examplesstarchatbeta}

\begin{figure}[htbp]
\hrulefill

\emph{Input:}

\hrulefill

\begin{lstlisting}[language=C++,breaklines=true,basicstyle=\ttfamily\scriptsize]
#include<stdio.h>
#include<math.h>
#include<vector>
#include<string>
#include<algorithm>
using namespace std;
#include<stdlib.h>
int specialFilter(vector<int> nums){
    int num=0;
    for (int i=0;i<nums.size();i++)
    if (nums[i]>10)
    {
        string w=to_string(nums[i]);
        if (w[0]%2==1 and w[w.length()-1]%2==1) num+=1;
    }
    return num;
}
\end{lstlisting}

Provide a concise natural language description of the code using at most 296 characters.

\hrulefill

\emph{StarChat-$\beta$:}

\hrulefill

This code is a solution to the "Special Array" problem on LeetCode. It uses C++ and includes several standard libraries, including vector, string, and algorithm.

The code starts by declaring some variables and importing the necessary libraries. It then defines a specialFilter function that takes

\hrulefill

\emph{Input (without previous input):}

\hrulefill

This code is a solution to the "Special Array" problem on LeetCode. It uses C++ and includes several standard libraries, including vector, string, and algorithm.

The code starts by declaring some variables and importing the necessary libraries. It then defines a specialFilter function that takes

Write functional code in C++ according to the description.
\begin{lstlisting}[language=C++,breaklines=true,basicstyle=\ttfamily\scriptsize]
#include<stdio.h>
#include<math.h>
#include<vector>
#include<string>
#include<algorithm>
using namespace std;
#include<stdlib.h>
int specialFilter(vector<int> nums){
\end{lstlisting}

\hrulefill

\emph{StarChat-$\beta$:}

\hrulefill

\begin{lstlisting}[language=C++,breaklines=true,basicstyle=\ttfamily\scriptsize]
    int n = nums.size();
    int sum = 0;
    for(int i = 0; i < n; i++){
        sum += pow(nums[i], 2);
    }
    int target = (int)sqrt(sum);
    vector<int> res;
    for(int i = 0; i < n; i++){
        if(nums[i] > target){
            res.push_back(nums[i]);
        }
    }
    return res;
}
\end{lstlisting}

\hrulefill
\caption{\textbf{Incorrect solution from StarChat-$\beta$ to a \evale{} C++ problem.} StarChat-$\beta$ has a bias for excessively long answers. StarChat-$\beta$'s description includes irrelevant parts like "..problem on LeetCode..includes several standard libraries.." (despite the problem not existing on LeetCode). As the description exceeds the 296-character limit for the code, it is cut off at the end.}
\label{fig:starchatbetacpp}
\end{figure}

\FloatBarrier

\subsection{Diff Codegen}
\label{sec:examplesdiffcodegen}

\begin{figure}[htbp]
\hrulefill

\emph{Input:}

\hrulefill

\begin{lstlisting}[language=Python,breaklines=true,basicstyle=\ttfamily\scriptsize]
<NME> truncate_number.py
<BEF> def truncate_number(number: float) -> float:
    """ Given a positive floating point number, it can be decomposed into
    and integer part (largest integer smaller than given number) and decimals
    (leftover part always smaller than 1).

    Return the decimal part of the number.
    >>> truncate_number(3.5)
    0.5
    """
    return number % 1.0 + 1.0
<MSG> Fix bugs in truncate_number.
<DFF>
\end{lstlisting}

\hrulefill

\emph{Diff Codegen 2B}

\hrulefill

\begin{lstlisting}[language=Python,breaklines=true,basicstyle=\ttfamily\scriptsize]
@@ -4,7 +4,7 @@ def truncate_number(number: float) -> float:
     and integer part (largest integer smaller than given number) and decimals
     (leftover part always smaller than 1).
 
-    Return the decimal part of the number.
+    Return
\end{lstlisting}

\hrulefill

\caption{\textbf{Incorrect solution from Diff Codegen 2B to a \evalfd{} Python problem.} Diff Codegen 2B suggests an irrelevant diff modifying parts of the docstring. The model commonly outputs diffs that modify the docstring or an import statement and rarely addresses the actual bug.}
\label{fig:diffcodegenjava}
\end{figure}

\FloatBarrier

\section{Limitations and Future Work}
\label{sec:limitations}

\paragraph{Model Execution} A promising avenue for improving performance on \evalf{} is letting the model execute the given code or its own generated code and inspect its output~\citep{chen2022codet,chen2023teaching,yasunaga2021break,li2022transrepair,gao2023pal,dong2023self,zhang2023coder,madaan2023self,ni2023lever,gou2023critic,hu2023code,taylor2022galactica,nye2021show}. This could allow the model to discover which unit tests are failing and for what reason. The model could then simply iterate on the function until all unit tests are passing. We leave explorations of this strategy to improve performance on \eval{} to future work.

\paragraph{Multi-file changes} For the creation of \data{}, we have filtered out any commits that affect multiple files to ensure commits are very specific and account for the fact that most current models are only capable of operating on a single file. Allowing models to take multiple files as input and modify multiple files given a single instruction is a promising direction for future work. There is active research on using repository-level context~\citep{ding2022cocomic,shrivastava2023repofusion,shrivastava2023repository,zhang2023repocoder,liu2023repobench} and the necessary long context windows~\citep{dai2019transformer,press2021train,Sun2021DoLL,dao2022flashattention,peng2023rwkv,liu2023lost,chen2023extending}.

\paragraph{Length-awareness} Current Code LLMs including \model{} struggle with awareness about the length of their generated output. For \evale{}, we instruct the models to limit their output to a given number of characters. While it is trivial for humans to count characters and adhere to the limit, all models tested frequently generate far too many characters. Prior work has shown that human raters are biased towards preferring longer texts~\citep{wu2023style} regardless of content. All models evaluated are instruction tuned on text that was at least indirectly assessed by human raters, hence they may be biased towards generating longer texts even if it means including literary bloat.

\paragraph{Better evaluation} Evaluating code instruction models is challenging for several reasons: \textbf{(1)~Prompting:} The prompt can significantly impact the performance of large language models~\citep{brown2020language,zhou2022large,muennighoff2022sgpt,babe2023studenteval}. To ensure fair evaluation we use the prompting format put forth by the respective authors of the models and a simple intuitive prompt for models without a canonical prompt (see \autoref{sec:prompts}). However, this may put models without a canonical prompt recommendation (e.g. BLOOMZ, GPT-4) at a slight disadvantage. \model{} and \modelx{} perform best when prompted using the same format we use during training (\autoref{fig:octocoder}) and we recommend always using this format at inference.\\
\textbf{(2)~Processing:} Models may accidentally impair otherwise correct code by e.g. including a natural language explanation in their output. We largely circumvent this issue through the use of strict stopping criteria and careful postprocessing (e.g. for GPT-4 we check if it has enclosed the code in backticks, and if so, extract only the inner part of the backticks discarding its explanations).\\
\textbf{(3)~Execution:} When executing code to compute pass@k, it is important that the generated code matches the installed programming language version. Models may inadvertently use expressions from a different version (e.g. they may use the Python 2 syntax of \verb|print "hi"|, which would fail in a Python 3 environment). In our evaluation, we did not find this to be a problem, however, as models become more capable, it may make sense to specify the version. Future prompts may include the version (e.g. ``use JDK 1.18.0'') or provide models with an execution environment that has the exact version installed that will be used for evaluation.\\
\textbf{(4)~Comprehensiveness:} Executing code can only reflect functional correctness lacking a comprehensive understanding of quality. Compared to execution-based evaluation, the human judgment of code quality can be considered more comprehensive as humans can consider factors beyond correctness. Directly hiring human annotators can be inefficient and expensive, and therefore researchers have explored approaches to automate human-aligned evaluation via LLMs~\citep{fu2023gptscore, liu2023gpteval, zhuo2023large}. However, recent work~\citep{wang2023far} suggests LLM-based evaluation can be biased towards certain contexts. Future work on automating the human-aligned evaluation of instruction tuned Code LLMs while avoiding such bias is needed.

\paragraph{Reward Models} Our commit datasets, \data{} and \dataft{}, also lend themselves well for learning human preferences~\citep{ethayarajh2024kto,rafailov2024direct}. The changed code after a commit generally represents a human-preferred version of the code (else the code would not have been modified). Thus, one could train a reward model that given the code before and after a commit, learns that the code afterward is better. Similar to prior work~\citep{christiano2017deep,ouyang2022training}, this reward model could then be used to guide a language model to generate code that is preferred by humans.


\section{Version Control}
\label{sec:version}

\textbf{V1 → V2:}
\begin{itemize}
    \item Added \autoref{sec:explainmetrics} on \evale{} metrics
    \item Added \autoref{sec:explainfim} on Fill-in-the-Middle
    \item Expanded the motivation for limiting the number of tokens in \dataft{} in \autoref{sec:compare}
    \item Fixed the StarCoder \evalfd{} in \autoref{sec:docstrings} thanks to Abhijeet Awasthi 
    \item Specified the programming languages of CommitPackFT (CommitPack has 350 languages while CommitPackFT has 277 after filtering)
    \item Made small writing improvements throughout
\end{itemize}


\FloatBarrier
\newpage

\section{\badpack{}}
\label{sec:octobadpack}

\begin{figure*}[htbp]
    \centering
    {{\includegraphics[width=0.45\textwidth]{figures/octopack.png}}}
    \qquad
    {{\includegraphics[width=0.45\textwidth]{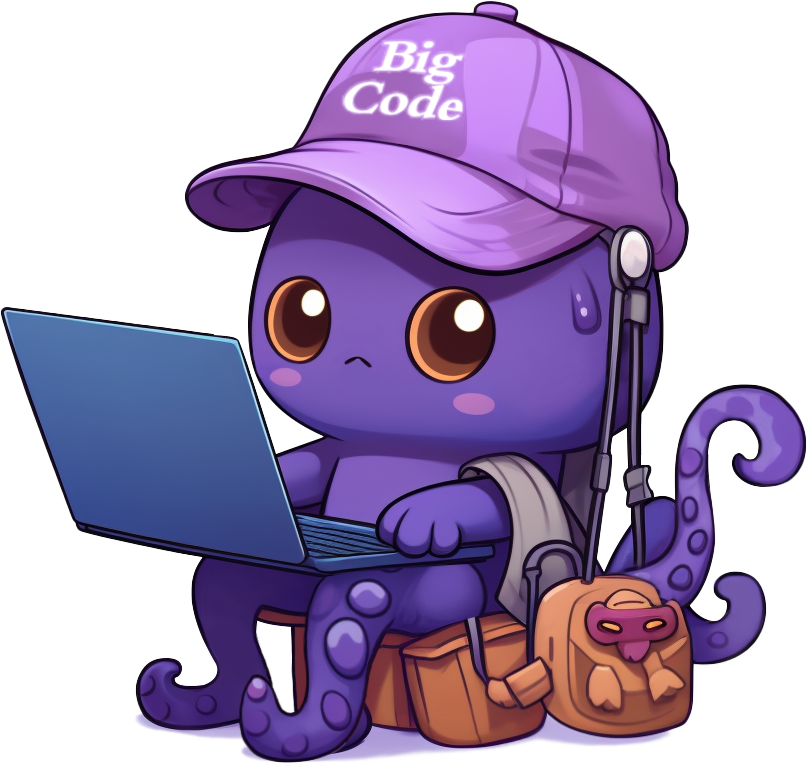}}}
    \caption{\pack{} (left) and her evil brother \badpack{} (right).}
    \label{fig:cdp}
\end{figure*}

\end{document}